\documentclass[review]{elsarticle}

\journal{Journal of \LaTeX\ Templates}

\biboptions{authoryear}
\usepackage[utf8]{inputenc} % allow utf-8 input
\usepackage[T1]{fontenc}    % use 8-bit T1 fonts
\usepackage{url}            % simple URL typesetting
\usepackage{booktabs}       % professional-quality tables
\usepackage{amsfonts}       % blackboard math symbols
\usepackage{nicefrac}       % compact symbols for 1/2, etc.
\usepackage{microtype}      % microtypography
\usepackage{amsfonts,amsmath,amssymb,amsthm,bm}
\usepackage{algorithm}
\usepackage{algorithmic}
\usepackage{graphicx}
\usepackage{booktabs}
\usepackage{makecell}
\usepackage{color}

\theoremstyle{plain}

\newcommand{\RR}{\mathbb{R}}                                     % real numbers
\newcommand{\bi}{\begin{itemize}}
\newcommand{\ei}{\end{itemize}}
\newcommand{\ba}{\begin{array}}
\newcommand{\ea}{\end{array}}

\DeclareMathOperator*{\argmax}{arg\,max}

\newcommand{\pivec}{\vec{\pi}}
\newcommand{\ovec}{\vec{o}}

\newcommand{\avec}{\vec{a}}
\newcommand{\E}{\mathbb{E}}

\newcommand{\brcksq}[1]{\left[#1\right]}

\newcommand{\eq}[1]{\begin{align}#1\end{align}}

\newcommand{\eps}{\epsilon}

\begin{document}

\begin{frontmatter}

\title{Decentralized policy learning with partial observation and mechanical constraints for multiperson modeling}





\author[1,2,3]{Keisuke Fujii\corref{mycorrespondingauthor}} 
\cortext[mycorrespondingauthor]{Corresponding author}
\ead{fujii@i.nagoya-u.ac.jp}
\author[4,2]{Naoya Takeishi}
\author[5,2]{Yoshinobu Kawahara}
\author[1]{Kazuya Takeda}

\address[1]{Graduate School of Informatics, Nagoya University, Nagoya, Aichi, Japan} 
\address[2]{Center for Advanced Intelligence Project, RIKEN, Osaka,Japan} 
\address[3]{PRESTO, Japan Science and Technology Agency, Tokyo, Japan} 
\address[4]{\textcolor{black}{Graduate School of Engineering, The University of Tokyo, Tokyo, Japan}} 
\address[5]{Graduate School of Information Science and Technology, Osaka University, Osaka,  Japan} 

\begin{abstract} % less than 250 words
Extracting the rules of real-world multi-agent behaviors is a current challenge in various scientific and engineering fields. 
Biological agents independently have limited observation and mechanical constraints; however, most of the conventional data-driven models ignore such assumptions, resulting in lack of biological plausibility and model interpretability for behavioral analyses. 
Here we propose sequential generative models with partial observation and mechanical constraints in a decentralized manner, 
which can model agents' cognition and body dynamics,
and predict biologically plausible behaviors. 
We formulate this as a decentralized multi-agent imitation-learning problem, leveraging binary partial observation and decentralized policy models based on hierarchical variational recurrent neural networks with physical and biomechanical penalties. 
Using real-world basketball and soccer datasets, we show the effectiveness of our method in terms of the constraint violations, long-term trajectory prediction, and partial observation.
Our approach can be used as a multi-agent simulator to generate realistic trajectories using real-world data.
\end{abstract}


\end{frontmatter}

%\linenumbers

\section{Introduction}
\label{sec:introduction}
Extracting the rules of biological multi-agent behaviors in complex real-world environments from data is a fundamental problem in a variety of scientific and engineering fields.
For example, animals, vehicles, pedestrians, and athletes observe others' states and execute their own actions with body constraints in complex situations.
In these processes, they observe others' movements and make decisions, for example, based on their experiences and knowledge, under spatiotemporal constraints. 
Pioneering studies proposed various ways of rule-based modeling such as in human pedestrians \citep{Helbing95} and animal groups \citep{Couzin02} using hand-crafted functions (e.g., social forces).
Recent advances in machine learning have enabled data-driven modeling of such behaviors (see Section~\ref{sec:related}).
These problems are formulated as imitation learning \citep{Ross11,Le17}, generative adversarial learning \citep{Chen18generating,Hsieh19}, or (simply) sequential learning, such as leveraging sequential generative models \citep{Chung15,Fraccaro16,Karl17,Fraccaro17}.

Most previous studies on data-driven models used the following three concepts to improve prediction performance: (1) they can fully utilize environmental information; % or that based on pre-determined rules (e.g., schooling fish models \cite{Couzin02}) 
(2) they can optimize communication on the basis of the centralized control; and (3) they ignore the mechanical constraints of the agent's body, resulting in biologically unrealistic behaviors.
Although such ideas improve the predictability, 
interpretable modeling based on biological plausibility \citep{Cichy19} is critical for scientific understanding, which is the motivation for this article.
For the first and second assumptions, organisms in the real-world independently have limited communication and observation. 
In this case, they should be considered as partially observable decentralized systems for analysis of their observations. 
For the third assumption, the body constraints such as inertia during motion planning can be described by the principles of robotics and computational neuroscience \citep{Flash85,Uno89}.
Therefore, the decentralized modeling of the agent's observation and of biologically plausible motions would contribute to the understanding of real-world multi-agent behaviors, which is one of the challenges, for example, in biological science.

In this article, we propose sequential generative models with partial observation and mechanical constraints in a decentralized manner, which visualize whose information the agents utilize and predict biologically plausible long-term behaviors. 
We formulate this as a decentralized multi-agent imitation-learning problem, leveraging explicit partial observation models with a Gumbel-softmax reparameterization \citep{Jang17,Maddison17} (see Sections~\ref{ssec:gumbel} and \ref{ssec:observation}) 
and decentralized policy models based on hierarchical sequential generative models with stochastic latent variables and mechanical constraints (see Sections~\ref{ssec:vrnn}, \ref{ssec:biomech}, and~\ref{ssec:policy}).
A team sport is an example that can be addressed with the above approach. Players observe others' states \citep{Fujii15a,Fujii15b}, % 
while actively and/or passively ignoring less informative agents \citep{Fujii16} regardless of the distance, and execute complex actions.
We investigate the empirical performance by using real-world ball game datasets. 

In summary, our main contributions are as follows: 
(1) We propose decentralized policy models with interpretable partial observation to predict biologically plausible long-term behaviors, contributing to the analysis and understanding of real-world decentralized multi-agent behaviors;
(2) the technical novelties of our policy models are explicit binary partial observation, decentralized modeling, and mechanical constraints, which can be compatible with many existing deep generative models;
(3) our approach is validated by our visualizing, evaluating, and counterfactually manipulating partial observation and predicted plausible behaviors using real-world basketball and soccer datasets.
Note that a purely rule-based approach can explicitly define and discuss the functions to imitate behaviors, but sometimes has difficulty in modeling numerous observation, decision-making, and movement-related functions of more complex behaviors such as in team sports. %  by utilizing currently available trajectory data
Thus we consider that, in our problems, such a data-driven approach is a better choice than a purely rule-based approach. 
We regard here ``biological plausibility'' as decentralized learning and soft constraints (e.g., represented as penalty functions) rather than hard constraints that rigorously satisfy the requirements (e.g., equations of motion) because the governing rules of our problem are largely unknown (otherwise a rule-based approach would be more effective).
In addition, we  consider here ``interpretability'' as an approach that visualizes whose information the agents utilize for nonlinear neural network models, rather than input feature analysis or an inherently explainable approach such as linear models and decision trees.
% ; and (4) our approach based on machine learning will bridge various behavioral and computer sciences, because we applied a data-driven modeling to real-world multi-agent behaviors based on the principles of robotics and computational neuroscience (e.g., POMDP: partially observable Markov decision process \cite{Bernstein02,Kaelbling98} and mechanical constraints in Section \ref{sec:background}).
In the remainder of this article, we describe the background of our problem and our method in Sections~\ref{sec:background} and~\ref{sec:proposed}, review related studies in Section~\ref{sec:related}, and present experimental results and conclusions in Sections~\ref{sec:experiments} and~\ref{sec:conclusion}.
 
% ==========

\section{Background}
\label{sec:background}

We formulate our problem as a decentralized multi-agent imitation-learning problem in a partially observable Markov decision process (POMDP) to analyze their observations, with the assumption that multiple agents in team sports may behave in a decentralized manner in a certain period (e.g., several seconds) because players have limited resources (e.g., time and observation) to decide their appropriate actions. 
Here, we introduce the background of elements in our model: imitation learning for a decentralized POMDP, observation models with an attention mechanism, sequential generative models for policy modeling, and biomechanical constraints. 

\subsection{Imitation learning in a decentralized POMDP}
%\vspace{-1pt}
\label{ssec:pomdp}
Here we consider a multi-agent problem that can be formulated as a decentralized POMDP \citep{Bernstein02,Kaelbling98,Mao19,Amato19}. % 
It is defined as a tuple $(K, S, A, \mathcal{T}, R, O, Z, \gamma)$, 
where $K$ is the fixed number of agents,;
$S$ is the set of states $s$;
$A=[A_1, \ldots, A_K]$ represents the set of joint action $\avec \in A$ (for a variable number of agents), with $A_k$ being the set of local action $a_k$ that agent $k$ can take;
$\mathcal{T}(s'| s,\avec): S \times A \times S \rightarrow [0,1]$ is the transition model;
$R=[R_1, \ldots, R_K]: S \times A \rightarrow \RR^K$ is the joint reward function;
$O = [O_1, \ldots, O_K]$ is the set of joint observation
$\ovec \in O$ controlled by the observation function $Z: S \times A \rightarrow O$; and $\gamma \in [0,1]$ is the discount factor. % 
In on-policy reinforcement learning, the agent learns a policy $\pi_{k}: O_k \times A \rightarrow [0,1]$
%$\pi_{k}: S \times O_k \rightarrow A_k$ 
that can maximize $\mathbb{E}[G_k]$, where $G_k = \sum_{t=1}^{T} \gamma^{t} R_{k}^{t}$ is the discounted return and $T$ is the time horizon. 
In a decentralized multi-agent system in complex real-world environments (e.g., team sports), transition and reward functions are difficult to design explicitly. % model generally
Instead, if we can utilize the demonstrations of expert behaviors (e.g., trajectories of professional sports players), we can formulate and solve our problem as imitation learning (i.e., as a machine-learning problem).
In other words, if the problem satisfies the two conditions, imitation learning is one of the better options.
Here, for example, in team sports, the state includes all the position, velocity, and acceleration of all players. The actions include each player's next velocity or acceleration, and the observations include information for a limited number of other players.

% \subsection{Multi-agent Imitation Learning in POMDP}
The goal in imitation learning is to learn a policy $\pi$ that imitates an expert policy $\pi_E$ given demonstrations from that expert \citep{schaal1996learning,Ross11}.
In multi-agent imitation learning, we have $K$ agents to achieve a common goal. 
On the basis of the notation in \citep{Ross11}, in the case of a fully centralized multi-agent policy for clarity, let $\pivec(\ovec) := \avec$ denote the joint policy that maps the joint observation $\ovec = [o_1,\ldots,o_K]$ into $K$ actions $\avec = [a_1,\ldots,a_K]$. 
Training data $\mathcal{D}$ consists of multiple demonstrations of $K$ agents.  
The decentralized setting decomposes the joint policy $\pivec = [\pi_1,\ldots,\pi_K]$ into $K$ policies, sometimes tailored to each specific agent index or role.
The loss function is then
$\mathcal{L}_{imitation} = \sum_{k=1}^K\E_{s_k \sim d_{\pi_k}}\left[ \ell(\pi_k(o_k)) \right],$
where $d_{\pi_k}$ is the distribution of states experienced by joint policy $\pi_k$ ($o_k$ is determined by $s_k$ and $a_k$) and $\ell$ is the imitation loss defined over the demonstrations.
This formulation assumes the assignment of appropriate roles for each agent such as solving a linear assignment problem \citep{Papadimitriou82} (see Appendix~B). % \ref{app:roleassign}).

%\vspace{-2pt}
\subsection{Approaches for partial observation models}
% using Gumbel-Softmax reparameterization
\label{ssec:gumbel}
%\vspace{-1pt}
Partial observation in the context of the POMDP typically indicates that the agent cannot directly observe the underlying state.
However, in real multiperson systems, people sometimes actively ignore less informative others. 
Since these two cases cannot be distinguished from the  data obtained, we formulate both cases as the former POMDP. 
Here we introduce several approaches for neural network--based partial observation using soft and hard attention mechanisms.
The details and other approaches are described in Appendix~A. %\ref{app:observation}.

Attention is widely used in % applications including
natural language processing \citep{Bahdanau14}, computer vision \citep{Wang18}, and multi-agent reinforcement learning (MARL) in virtual environments \citep{Mao19,Jiang18,Iqbal19}. 
Soft attention calculates an importance distribution of elements (e.g., agents).
%: $w_{k}=\exp(p_k) / \sum_{i=1}^{K}\exp(p_i)$, where $p_k$ is a scalar variable for an agent $k$.
Soft attention is fully differentiable and thus can be trained with back-propagation.
However, it usually assigns nonzero probabilities to unrelated elements; i.e., it cannot directly reduce the number of agents.

Hard attention focuses solely on an important element but is basically nondifferentiable because of the selection based on sampling.
Among differentiable models with discrete variables, an approach using Gumbel-softmax reparameterization (see, e.g., \citet{Jang17}) can be effectively implemented via a continuous relaxation of a discrete categorical distribution. % Maddison17
%In summary, given $K$-categorical distribution parameters $p$, a differentiable $K$-dimensional one-hot encoding sample $G$ from the Gumbel-Softmax distribution can be computed as:
%\begin{align*}
%$G(\text{log} p)_k = \text{exp}((\text{log} p_k + \eps)/\tau ) / \sum_{i=0}^K \text{exp}((\text{log} p_i + \eps)/\tau ) ,$ % \frac{
%\end{align*}
%where $\eps$ are i.i.d. samples from a Gumbel$(0,1)$ distribution, i.e., 
%$\eps = -\text{log}(-\text{log}(u)), \; \; u \sim \mathcal{U}[0,1]$.
%$\mathcal{U}$ is a uniform distribution and $\tau$ is the softmax temperature parameter. 
However, the resulting one-hot vector is insufficient to represent a multi-hot observation. % $[G(\text{log} p)_1,\ldots,G(\text{log} p)_K]$
Our model enables a multi-hot representation as the observation in multi-agent \textcolor{black}{systems for each agent}.
Other approaches, such as relational inference (see, e.g., \cite{Kipf18}), can learn multi-agent interactions by learning graph structures (see also Appendix~A). %\ref{app:related}). 
Since we aim to model decentralized long-term multi-agent behaviors, we adopt an approach to explicitly train independent dynamical models of each agent. 
 
%\vspace{-5pt}
\subsection{Sequential generative models}
\label{ssec:vrnn}
%\vspace{-1pt}
Agent trajectories or actions in the real world
have been modeled recently as sequential generative models (see also Section~\ref{sec:related}).
Here we consider single-agent modeling for simplicity.
Let $a_{\leq T} = \{ a_1, \dots, a_T \}$ denote a sequence of actions of length $T$. 
The goal of sequential generative modeling is to learn the distribution over sequential data $\mathcal{D}$ consisting of multiple demonstrations. 
We assume that all sequences have the same length $T$, but in general, this does not need to be the case.
A common approach is to factorize the joint distribution and then maximize the log-likelihood 
%
%\begin{align}
$\theta^* = \argmax_{\theta} \sum_{a_{\leq T} \in \mathcal D} \log p_{\theta} (a_{\leq T}) = \argmax_{\theta} \sum_{a_{\leq T} \in \mathcal D} \sum_{t=1}^T \log p_{\theta} (a_t | a_{<t}),$
%label{eq:condprobs}
%end{align}
%
where $\theta$ denotes the model's learnable parameters, such as recurrent neural networks (RNNs).

%
%\textbf{Stochastic latent variable models.}
%
However, RNNs with simple output distributions often struggle to capture highly variable and structured sequential data (e.g., multimodal behaviors) \citep{Zhan19}. 
Recent work in sequential generative models addressed this issue by injecting stochastic latent variables into the model and optimization using amortized variational inference to learn the latent variables (see, e.g., \cite{Chung15,Fraccaro16,Goyal17}); see Section~\ref{sec:related}. % 
Among these methods, a variational RNN (VRNN) \citep{Chung15} has been widely used in base models for multi-agent trajectories \citep{Yeh19,Zhan19} with unknown governing equations (see Section~\ref{ssec:overview}).
Note that our VRNN-based approach is compatible with other sequential generative models.
A VRNN is essentially a variational autoencoder (VAE) \citep{Kingma14} conditioned on the hidden state of an RNN and is trained by maximization of the (sequential) evidence lower bound (ELBO):
\vspace{-3pt}
\begin{align}
\mathbb{E}_{q_{\phi}(z_{\leq T} \mid a_{\leq T})} &
\Bigg[ \sum_{t=1}^T \log p_{\theta}(a_t \mid z_{\leq t}, a_{<t}) \label{eq:vrnn_elbo} 
\\&-D_{KL} \Big( q_{\phi}(z_t \mid a_{\leq t}, z_{<t}) || p_{\theta}(z_t \mid a_{<t}, z_{<t}) \Big) \Bigg], \nonumber
\vspace{-3pt}
\end{align}
where $z$ is a stochastic latent variable. 
The first term is the reconstruction term and $p_{\theta}(a_t \mid z_{\leq t}, a_{<t})$ is the generative model.
The second term is the Kullback-Leibler (KL) divergence between the approximate posterior or inference model $q_{\phi}(z_t \mid a_{\leq t}, z_{<t})$ and the prior $p_{\theta}(z_t \mid a_{<t}, z_{<t})$.
Eq.~(\ref{eq:vrnn_elbo}) can be interpreted as the ELBO of a VAE summed over each timestep $t$ (see also Appendix~C). % \ref{app:sequential}).

\subsection{Biomechanical constraints} 
\label{ssec:biomech}
%\vspace{-2pt}
In robotics and computational neuroscience, the path planning in smooth and flexible biological motions is a classical problem (see, e.g., \cite{Flash85}). %Uno89}. 
Most research has focused on individual multijoint motions.
% However, in our multi-agent cases,
Since multi-agent datasets often include only a mass point for each agent,
we can use the principles of smooth mass-point motions. % such as end-point effectors.
A minimum-jerk principle \citep{Flash85} is a simple and well-known principle for smooth motor planning of voluntary motion. % in an end-point effector.
It minimizes the motor cost: 
$C=(1/2)\int_0^T ||(d^3x/dt^3)||^2_2 dt$,
where $x$ is a multidimensional position vector and $\|\cdot\|_2$ is the Euclidean norm.
There is some research showing that this principle can be applied to human locomotion \citep{Pham07} and driving behavior \citep{ Dias20}. 
% Although we do not know the researches applying the principle to multi-agent sports, its application seems to be natural. 
We propose a biomechanical constraint for learning a multi-agent model inspired by this principle for the first time.
% simply applicable 

\section{Proposed method}
\label{sec:proposed}
\begin{figure*}[t]%{0.48\linewidth} % sub
\centering
\includegraphics[width=1\columnwidth]{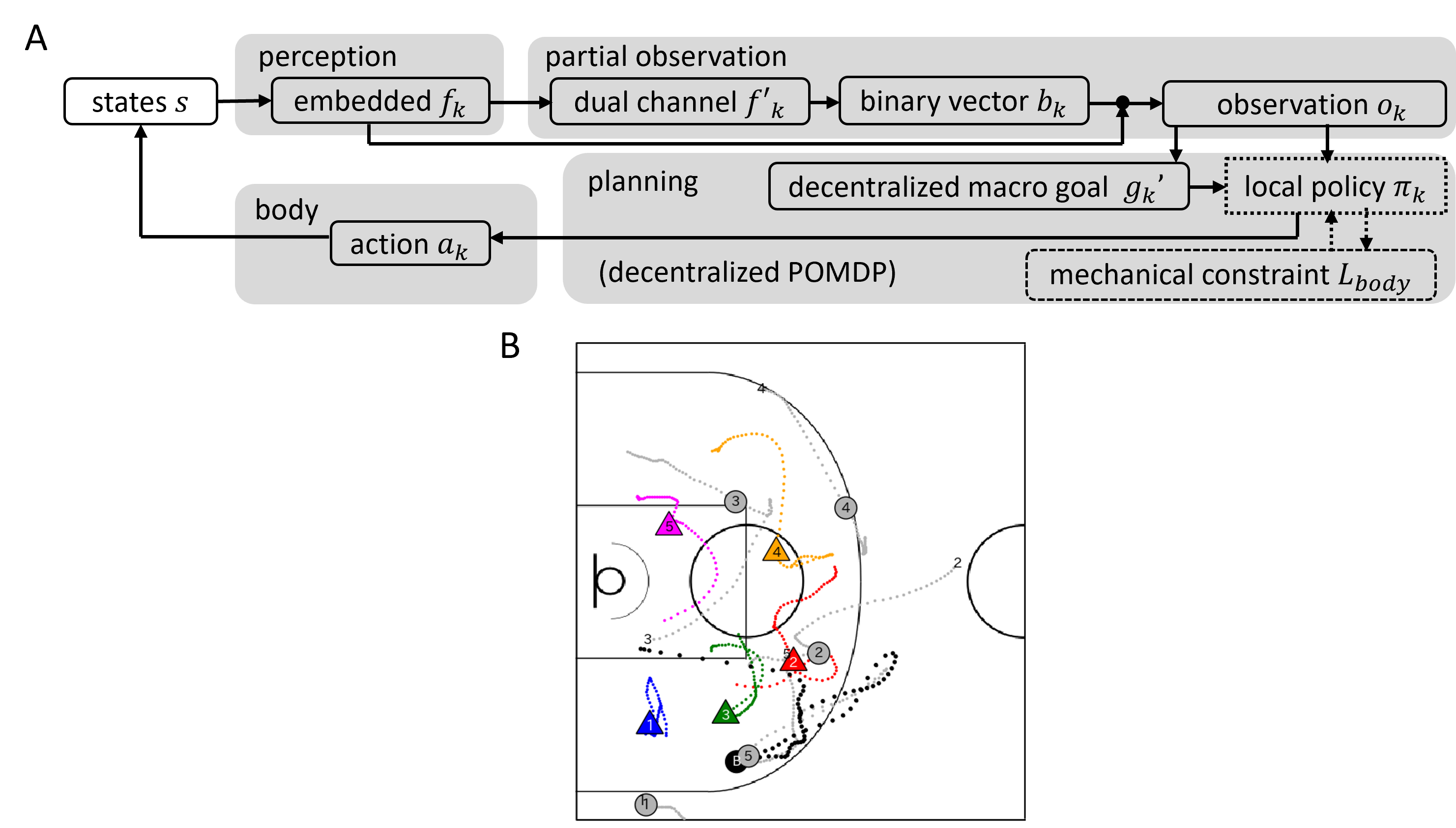}
\caption{{(A) Our model and (B) an example trajectory in basketball. In (A), f}or clarity, we omitted the time index $t$. 
The agent $k$ perceives the state $s_{t-1}$ and outputs the action $a_{t,k}$.
The detailed configuration is described in Section~3.
{Details are described in the main text. In (B), colored triangles, gray circles, and the black circle are defenders, attackers, and the ball, respectively. }
}
\label{fig:diagram}
\vspace{-0pt}
\end{figure*}

In this section, we propose decentralized policy models with interpretable observations to predict plausible long-term behaviors. %  sequential generative models for 
We leverage binary partial observation models and a hierarchical VRNN with mechanical constraints in a decentralized manner (Fig.~\ref{fig:diagram}A).
We first provide an overview of our model, then we propose a binary observation and decentralized policies with mechanical constraints, and finally we describe the learning method. 

\vspace{-4pt}
\subsection{Overview} 
\label{ssec:overview}
%\vspace{-2pt}
% In our problem setting, 
As an imitation-learning problem for decentralized multi-agent systems in a POMDP, we aim to learn policies $\pivec = [\pi_1,\ldots,\pi_K]$ that imitate expert policies $\pivec_E = [\pi_{E_1},\ldots,\pi_{E_K}]$ given the expert action sequences of $K$ agents $a_{\leq T} = \{ a_{\leq T, 1}, \dots, a_{\leq T,K} \}$ under biologically realistic constraints.
As shown in Fig.~\ref{fig:diagram}A, the agent perceives the state $s_{t-1}$ and performs path planning for the action $a_{t,k}$. 
In this work, to interpret our model as a decentralized multi-agent policy, we train independent models for each agent.
% since we consider the decentralized model, we train independent models for each agent.

Our model consists of a partial observation, local policy, macro goal, and mechanical constraints.
{According to the data flow, we first describe a perception and} a binary vector $b_{t,k}$ of the partial observation model defined in Section~\ref{ssec:observation}. 
For example, we consider that, in team sports, the players perceive other players' information and obtain an abstract representation (as an embedded vector) and selectively choose information for a limited number of players (as the partial observation).
{Next, t}he model also uses macro goals \citep{Zhan19} {as auxiliary information (from a pretrained model)} for long-term prediction modified for our decentralized and partially observable problem (see Appendix~D). % \ref{app:macro_goal}).
Note that human behaviors are generated by a {policy function including tactical movements} for which rules are partially unknown \citep{Fujii19c} {as shown in Fig.~\ref{fig:diagram}B}. %  including cognitive and motor factors 
% For example, some cognitive factors (often higher-level factors such as tactics) remain unknown.
We therefore adopt a nonlinear transition model (see Section~\ref{sec:related}) 
for the local policy $\pi_k$, specifically based on a VRNN \citep{Chung15}, which can generate a trajectory with multiple variations (see, e.g.,  \cite{Yeh19,Zhan19,Sun19}). % 
{With the obtained observation as an input, the macro goal and local policy are learned. The macro-goal model is pretrained before the policy learning, and for the policy learning, we use the macro goal as weak supervision \citep{Zhan19}.} 
For example, in team sports, players move, including the intention of where to move for a length of a few seconds (we model it as the macro goal) while adapting to the situation in the moment.
% Again, our approach is also applicable to other sequential models. % generative

{In addition, we consider mechanical constraints as a penalty of the objective function.}
In previous work (e.g., \cite{Zhan19,Yeh19}), the action generated by $\pi_k$ was usually the agent's position.
However, these studies reported difficulty in learning velocity and acceleration (e.g., frequent changes in direction of motion in basketball), which are critical to our problem {because of the complex movements as shown in Fig.~\ref{fig:diagram}B}. 
We aim to obtain policy models to generate biologically plausible actions in terms of position, velocity, acceleration, and their relations.
Our solution incorporates mechanical constraints {as a penalty function} into policy learning.
We describe the details for the policy model, macro goal, and mechanical constraints in Section~\ref{ssec:policy}.

Note that we entirely consider biologically plausible architectures and loss functions of the neural network, whereas we sometimes exploit existing modules based on previous work, which do not necessarily consider the biological plausibility and cannot often provide concrete examples such as in team sports.

%\vspace{-5pt}
\subsection{Binary partial observation model} % 
\label{ssec:observation}
%\vspace{-1pt}
To model partial observation in a multi-agent setting, we propose a binary partial observation model that can visualize whose information the agents utilize. 
In this work, this is represented by the binary vector $b_{t,k}$.
% using Gumbel-Softmax reparameterization.
As shown in Fig.~\ref{fig:diagram}A {(perception and partial observation modules)}, the model is trained to
map input state vector $s_t = [s_{t,1},\ldots,s_{t,K}] \in \RR^{d_s \times K}$ to a latent binary vector $b_{t,k}$ of an agent $k$ at each time $t$, where $d_s$ is the dimension of the state for each agent.
We consider a partially observable environment, where each agent $k$ receives a binary vector
$b_{t,k} := [e_{k,1}^t, e_{k,2}^t, \ldots, e_{k,K}^t] \in \{0,1\}^K$.
% \in \RR^K, e_{k,i}^t \in \{0,1\}$.
That is, $b_{t,k}$ consists of an arbitrary number of 0's and 1's.
In this article, $b_{t,k}$ for an agent $k$ is designed to represent the importance of all agents, used as the observation coefficients or interpretable partial observation described below, which can visualize whose information the agents utilize in a multi-agent system.
% In this paper, $b_{t,k}$ for an agent $k$ is designed to represent the importance of all agents, used as the observation coefficients described below.
 
To obtain the binarized or multi-hot differentiable vector $b_{t,k}$, we linearly project a state vector $s_{t,k}$ into embedded matrices $f_{t,k} \in \RR^{K \times d_e}$ {(perception module with the full observation)} and into dual channels $f'_{t,k} \in \RR^{K \times 2}$ {(partial observation module)}, where $d_e$ is the embedding dimension.
The former embedding can transform the state into a distributed representation, which can be directly weighted by $b_{t,k}$ as described below (e.g., Cartesian coordinates cannot be directly weighted in principle). 
The latter dual-channel projection inspired by \citep{Liu19unsupervised} allows each dimension in $f'_{t,k}$ (and finally in $b_{t,k}$) to represent multiple agents' importance (not limited to the one-hot approach in Section~\ref{ssec:gumbel}). 
We perform a categorical reparameterization trick with Gumbel-softmax \textcolor{black}{reparameterization} \citep{Jang17} on the second dimension of the dual channel $f'_{t,k}$ (rather than on the first dimension as used in the attention mechanisms). %  in \ref{ssec:gumbel}
Since the two channels are linked together by Gumbel-softmax \textcolor{black}{reparameterization}, it is sufficient to simply pick one of them.
That is, $e_{k,i}^t = [\operatorname{Gumbel-softmax}([f'_{t,k}]_i)]_1$, where $[\cdot]_i$ denotes $[\cdot]$'s $i$th element.
% Since the two channels are linked together by Gumbel-Softmax, simply picking one of them is sufficient, hence we select the first channel as $o_{t,k}$.
This method is differentiable and allows direct back-propagation in our framework. 

We then compute the observation vector $o_{t,k}$ as the input to the subsequent policy learning.
We concatenate the elementwise product of binary vector $b_{t,k}$ (observation coefficients) and the embedded matrices $f_{t,k}$ for all agents: 
$o_{t,k} = [e_{k,1}^t [f_{t,k}]_1, \ldots, e_{k,K}^t [f_{t,k}]_K] \in \RR^{d_e K}$.
This allows us to eliminate the information from unrelated agents, and to focus on only the important agents.

%\vspace{-2pt}
\subsection{Decentralized hierarchical VRNN with macro goals and mechanical constraints} 
\label{ssec:policy}

\begin{figure}
\vspace{-0pt}
\centering
\includegraphics[width=0.7\linewidth]{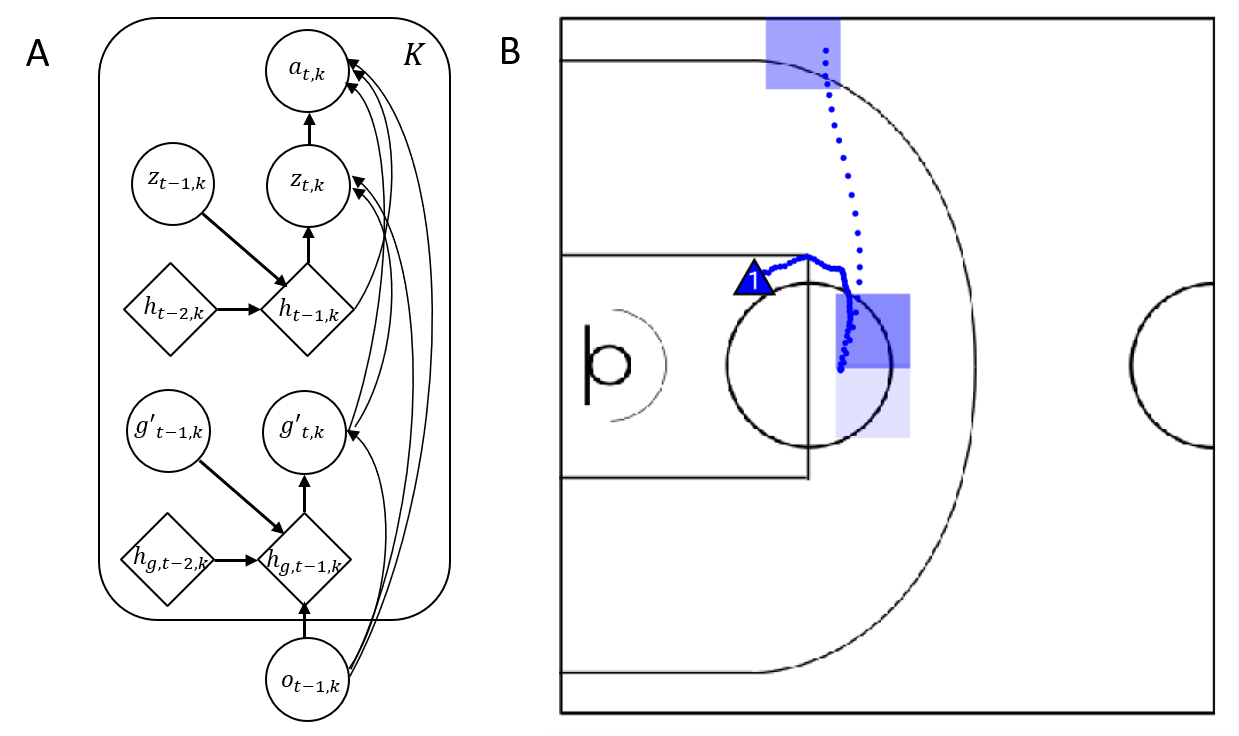}
\caption{Diagram and example of the components of our model. (A) Graphical illustration of the decentralized and hierarchical local policy model with macro goals (denoted as $g'$). Circles are stochastic variables and diamonds are deterministic variables. For specific variables, see the main text. (B) Example of a decentralized macro goal (boxes) for defender~1 in basketball. After reaching the macro goal in the center, the player moves toward the next macro goal at the top (middle).
The macro goals provide a compact summary of the players' sequences over a long time to encode long-term intent.
The decentralized macro goals are pretrained from the multi-agent trajectories and then used in the training of the policy model (A). For details, see Appendix~D.
\label{fig:graphical} }
\vspace{-0pt}
\end{figure}

%\noindent{\bf{Decentralized hierarchical VRNN with macro-goals.}}
First, we perform VRNN modeling with conditional context information including the observation and macro goals, 
as illustrated in Fig.~\ref{fig:graphical}A.
{Here we describe three important components: decentralized macro goals, local policy, and mechanical constraints.}

%\noindent{{\bf{Decentralized macro goals.}}}
\subsubsection{Decentralized macro goals}
To model an agent's macroscopic intent during path planning, we use weak labels for macro goals \citep{Zhan19} as illustrated in Fig.~\ref{fig:graphical}B.
{
The original macro goal or macro intent \citep{Zheng16,Zhan19}, obtained via some labeling functions, is defined as low-dimensional and spatiotemporal representations of the data for the learning of multi-agent long-term coordination.
Firstly, we label the macro goal programmatically via the labeling function to obtain macro goals; for example, the regions on the court in which players remain stationary in this case. The label (macro goal) is the one-hot encoding of the box that contains the position information. Then, the shared macro-goal model is learned via supervised learning of multilayer perceptrons by our maximizing the log-likelihood of macro-goal labels. The input of the model is the partial observation $o_{t-1,k}$. 
}

This approach is currently one of the best methods for long-term prediction because it uses future positional information trained by weak labels.
% (for details, see Appendix \ref{app:macro_goal}).
Here we use a macro goal as a part of the planner and modify it in a decentralized and partially observable setting (for details, see Appendix~D). % \ref{app:macro_goal}).
We use the partial observation $o_{t-1,k}$ {as an input} and independently learn the decentralized macro goal $g'_{t,k}$ (not shared between agents) {before learning of the local policy (as a pretrained model).
% The macro-goals provide a compact summary of the players' sequences over a long time to encode long-term intent.
The decentralized macro goals are pretrained from the multi-agent trajectories and then used as the auxiliary information in the training of the policy model (see also Appendix~D).}

%\noindent{{\bf{Local policy.}}
\subsubsection{Local policy}
On the basis of the observation and the macro goal as inputs, overall, our VRNN-based model for the local policy becomes
\vspace{-2pt}
\eq{
p_{\theta_k}(a_{t,k} | o_{< t, k}) =
\mathcal{N}(a_{t,k} | \mu_{dec}^{t,k}, (\sigma_{dec}^{t,k})^2),
\label{eq:cvrnn}
\vspace{-2pt}
}
where \textcolor{black}{$\mu_{dec}^{t,k}$ and $ \sigma_{dec}^{t,k}$ are generated from $\varphi_{dec}^k(o_{t-1,k}, z_{t,k}, h_{t-1,k}, g'_{t,k})$}, $\varphi_{dec}^k$ is the Gaussian VRNN decoder,
$z_t^k$ is the VRNN latent variable, and $h_{t,k}$ is the hidden state of an RNN. % at time $t$.
To sample macro goals, we train another RNN model \citep{Zhan19}: % in a similar manner to that in 
%\eq{ \label{eq:macro_policy}
$p(g'_{t,k} | g'_{< t,k}) =
\mathcal{N}(g'_{t,k} | \mu_{g'}^{t,k}, (\sigma_{g'}^{t,k})^2)
% \varphi_{g_k}(h_{g,t-1,k}, o_{t-1,k}),
$,
%}
where \textcolor{black}{$\mu_{g'}^{t,k}$ and $ \sigma_{g'}^{t,k}$ are generated from $ \varphi_{g'_k}(h_{g,t-1,k}, o_{t-1,k})$,}  $\varphi_{g'_k}$ is the Gaussian macro-goal decoder
and $h_{g,t-1,k}$ summarizes the history of macro goals.

%\vspace{+2pt}
%\noindent {\bf{Mechanical constraints.}}
\subsubsection{Mechanical constraints}
Next, %we model mechanical constraints.
we propose two types of mechanical constraint {as penalties of the objective function} for training our model for the first time: a biomechanical constraint for the smooth path planning in Section~\ref{ssec:biomech} 
and physical constraints in multiple dimensions (i.e., position, velocity, and acceleration).
The detailed objective function is described in Section~\ref{ssec:learning}.
Both constraints are soft constraints by adding penalties, rather than hard constraints that rigorously satisfy the requirements.
This is because hard constraints such as numerical differentiation and integration derived more errors in our case where the data does not have a high spatiotemporal resolution (see also Appendix~G). %\ref{app:dimension}).
% The results of the ineffective models with hard constraints are shown in appendix \ref{app:dimension}.
% We aim to obtain policy models to generate biologically plausible actions in terms of the above dimensions.

For the biomechanical constraint, we consider a minimum-jerk principle \citep{Flash85} as mentioned in Section~\ref{ssec:biomech}.
However, this principle assumes that we know the start and end times of the movement.
We then consider the penalty for minimum change in acceleration, considering only the current and next acceleration, in an  way analogous to the minimum torque change principle in multijoint motion \citep{Uno89}.
We then propose the penalty that minimizes the difference between the predicted acceleration $\hat{a}_{acc,t}$ and the true acceleration $a_{acc,t+1}$ in the next step.
Simply, the Euclidean norm $\|a_{acc,t+1}-\hat{a}_{acc,t} \|_2^2 $ or, similarly, a negative log-likelihood (NLL) $- \log p_{\theta}(a_{acc,t+1} \mid z_{\leq t}, o_{<t}, g'_{<t})$ in a probabilistic model can be used.
% in a deterministic model 

Secondly, we propose physical constraints between multiple dimensions to consistently learn the relations. % hip
% Obviously, 
The model can generally learn the output dimension (e.g., position) but not the other dimensions (e.g., velocity).
By contrast, learning multiple dimensions
% (e.g., position and velocity) 
results in physically unrealistic and inconsistent predictions between the dimensions.
We then propose two physical penalties for the predicted and eliminated output dimensions.
For clarity, the following example uses velocity.
For the predicted dimension, we propose a penalty between the directly predicted velocity and the indirectly predicted velocity denoted by $\hat{a}_{vel,t}$ and $\tilde{a}_{vel,t} =  (\hat{a}_{pos,t}-\hat{a}_{pos,t-1})/\Delta t$, respectively, where $\hat{a}_{pos,t}$ is the directly predicted position and $\Delta t$ is a sampling interval.
For example, a KL divergence between the two distributions (\textcolor{black}{see the next subsection}) or the Euclidean norm can be proposed.
% $\|\hat{a}_{vel,t}-\tilde{a}_{vel,t}\|_2^2 $.
It can be also computed for acceleration, but not for position in principle.
For the eliminated dimension, we propose a penalty between the indirect prediction $\tilde{a}_{m,t}$ and true ${a}_{m,t}$, where $m = \{pos,vel,acc\}$; for example, an NLL between the distribution of $\tilde{a}_{m,t}$ and ${a}_{m,t}$ (\textcolor{black}{see the next subsection}) or the Euclidean norm.
% (for the verification, see Appendix \ref{app:mechanics}). 
% We verified various constraints in Appendix \ref{app:mechanics}.
%  $\|a_{acc,t+1}-a_{acc,t} \|_2^2 $.
%\vspace{-2pt}
\subsection{Learning}
\label{ssec:learning}
%\vspace{-1pt}
The objective function becomes a sequential ELBO of the hierarchical VRNN and the penalties of the mechanical constraints for the probabilistic model.
The ELBO for agent $k$ is $\mathcal{L}_{vrnn} = $
%
%\begin{align}
%\small{
$ \mathbb{E}_{q_{\phi}(z_{\leq T} \mid o_{\leq t},g'_{\leq t})}  \sum_{t=1}^T 
[\log p_{\theta}(a_t \mid z_{\leq t}, o_{<t}, g'_{\leq t}) %\notag\\
- D_{KL} 
( q_{\phi}(z_t \mid o_{\leq t}, z_{<t}, g'_{\leq t}) || p_{\theta}(z_t \mid o_{<t}, z_{<t}, g'_{\leq t}) 
) ].$
%}
%\label{eq:cvrnn_elbo}
%\end{align}
% \normalsize{}
For brevity, the symbol of the agent $k$ is not shown here. 
The penalties for mechanical constraints $\mathcal{L}_{body}$ are
\vspace{-2pt}
\begin{align}
\sum_{t=2}^T \Big[&
\small{D_{KL} ( p_{\theta}(\hat{a}_{m',t} \mid z_{\leq t}, o_{<t}, g'_{\leq t}) || p_{\theta}(\tilde{a}_{m',t} \mid z_{\leq t}, o_{<t}, g'_{\leq t}) )} \nonumber
\\& - \log p_{\theta}(a_{acc,t+1} \mid z_{\leq t}, o_{<t}, g'_{\leq t}) \Big],
\label{eq:penalty}
\vspace{0pt}
\end{align}
% \notag\\
% 
where $m' = \{vel,acc\}$ indicates velocity and acceleration as action outputs, depending on experimental conditions.
The first term is the penalty for consistently learning the relationship between the directly and indirectly estimated dimensions.
If we eliminate some dimensions (e.g., position), we can add NLLs as the penalty between the distribution of the indirect prediction and the ground truth: $- \log p_{\theta}(a_{m,t} \mid z_{\leq t}, o_{<t}, g'_{<t})$.
The second term is the biomechanical smoothing penalty, which minimizes the difference between the distribution of the directly or indirectly predicted acceleration ($\hat{a}_{acc,t}$ or $\tilde{a}_{acc,t}$) and the observed next acceleration $a_{acc,t+1}$. % in the next step.
We jointly maximize $\mathcal{L}_{vrnn} + \lambda \mathcal{L}_{body}$ with respect to all the model parameters (for the specific regularization parameter $\lambda$, see Appendix~E). % \ref{app:training}). %  

%\vspace{-0pt}
\section{Related work}
\label{sec:related}

%\noindent{\bf{Deep generative models for sequential data.}}
\subsection{Deep generative models for sequential data}
There has recently been  increasing interest in deep generative models for sequential data, because of the flexibility of deep learning and (often probabilistic) generative models.
In particular, many researchers intensively developed methods for physical systems with governing equations, such as a bouncing ball and a pendulum.
These studies modeled latent linear state space dynamics (e.g., \cite{Karl17,Fraccaro17}) or ordinary differential equations (e.g., \cite{Chen18neural}) mainly without RNNs. %,Yildiz19

% as mentioned above, 
In biological systems with partially unknown governing equations (e.g., external forces and/or nontrivial interactions), RNN-based models are still used (see below). % the paragraph
Among the RNN-based generative models (e.g., \cite{Chung15,Fraccaro16}), we incorporate a VRNN \citep{Chung15} with a stochastic latent state into the observation model and prior knowledge about mechanics. % ,Goyal17

%\vspace{2pt}
%\noindent {\bf{Multi-agent trajectory prediction.}}
\subsection{Multi-agent trajectory prediction}
In existing research on various biological multi-agent trajectories, pedestrian prediction problems including rule-based models (see, e.g., \cite{Helbing95}) have been widely investigated for a long time.
Recent studies using RNN-based models (e.g., \cite{Alahi16,Gupta18}) effectively aggregated information across multiple persons using specialized pooling modules, but most of which predicted trajectories in a centralized manner. 
Vehicle trajectory prediction has been intensively researched, often with use of deep generative models based on RNNs \citep{Bansal18,Rhinehart19,Tang19}, whereas little research has focused on animals \citep{Eyjolfsdottir17,Johnson16, tsutsui2021flexible}.
For sports multi-agent trajectory prediction such as in basketball and soccer, most methods have leveraged RNNs \citep{Zheng16,Le17,Ivanovic18,Teranishi2020}, % Liu19naomi,
including VRNNs \citep{Zhan19,Yeh19,teranishi2022evaluation,fujii2022estimating}, although some have used generative adversarial networks (e.g., \cite{Chen18generating,Hsieh19}) without RNNs (see also the review in \cite{fujii2021data}). %,
Most of these studies assumed full observation to achieve long-term prediction in a centralized manner (e.g., \cite{Zhan19,Yeh19}), except for an image-based study on partial observation \citep{Sun19}.
% centralized control 
In contrast, we aim to model decentralized biological multi-agent systems and visualize their partial observations for behavioral analyses in real-world agents.

%\vspace{2pt}
%\noindent {\bf{Observation in multi-agent systems.}}
\subsection{Observation in multi-agent systems}
In purely rule-based models, researchers investigating animals and vehicles conventionally proposed predefined observation rules based on specific distances and visual angles (e.g., \cite{Couzin02}), the specific number of the nearest agents (e.g., \cite{Ballerini08}), or other environments (e.g., \cite{Yoshihara17}). 
However, it is difficult to define interactions between agents using predefined rules in general large-scale multi-agent systems. Thus,
methods for learning observation of agents have been proposed for MARL in virtual environments \citep{Mao19,Jiang18,Iqbal19,Liu20} and for real-world multi-agent systems \citep{Hoshen17,Leurent19,Li20,Fujii20}. %Teng19
We describe the soft and hard attention approaches we used in Appendix~A, whereas other approaches, such as relational or causal inference, can learn multi-agent interactions by learning graph structures (e.g., \cite{Kipf18,Graber20, Lowe20}) or sparse weights of the first layer (e.g., \cite{Tank18,Khanna19,fujii2021learning}). 
These methods include a physically interpretable approach \citep{Fujii19b,Fujii20} 
that can learn interactions especially in physical particles or oscillators.
% In real-world multi-agent systems, a few applications in vehicles \citep{Leurent19} and physical and biological systems \citep{Li20,Sun19} existed.
Instead, our approach aims to model decentralized smooth multi-agent systems that can also predict long-term behaviors.

%\vspace{-5pt}
\section{Experiments}
\label{sec:experiments}
%\vspace{-2pt}
We quantitatively compared our models with various baselines using basketball and soccer game datasets.
The former includes observation of smaller players and more frequent acceleration due to the smaller playing area than for the latter (we thus performed some detailed analyses with the former dataset). % fine-grained 
% Due to the difficulty in evaluating a generative model with a single criterion \citep{Yeh19}, we evaluated several different metrics.
{Again, note that we propose a data-driven model with new neural network architectures and penalty functions, rather than directly modeling domain knowledge as explicit formulae (e.g., equations of motion). 
Therefore, w}e validated the effectiveness of our mechanical constraints (penalties) with various constraint losses.
We hypothesize that our methods show greater consistency between the directly and indirectly estimated dimensions (e.g., velocity and acceleration), and more smoothness in acceleration.
Next, we validated our partial observation model by visualizing and quantitatively analyzing the estimated observations. 
For this analysis, since we have no ground truth of the observation, we validated our model  on the basis of the domain knowledge.
Moreover, we investigated the long-term trajectory prediction performances as a validation.
{Since improvement of the prediction performance is not our purpose, we hypothesize our methods show performances similar to those of conventional methods}. 

% We mainly visualize the results of the basketball dataset, but quantitatively validated our methods using both datasets to demonstrate versatility. 
% In the experiments, 
We focused on learning team defense policies (see, e.g., \cite{Le17}) because they can be regarded as decentralized agents for a certain period (several seconds).
We provided the states of the offense players and the ball as conditional inputs to our models, and updated the states using methods in  \citep{Le17}.
The code is available at \url{https://github.com/keisuke198619/PO-MC-DHVRNN}.
% Modeling the ball and offense was left for future work.

%\vspace{-4pt}
\subsection{Common setups}
%\vspace{-1pt}
%\noindent{\bf{Baselines.}}
\subsubsection{Baselines}
We compared our approach with four baselines: (1) Velocity, (2) RNN-Gauss, (3) VRNN, and (4) VRNN-macro. The first two models are used as a sanity check for evaluating the trajectory prediction performance. 
First, we used velocity extrapolation as a simple baseline; i.e., each agent prediction was linearly extrapolated from its previously observed velocity.
The second baseline is an RNN implemented with use of a gated recurrent unit % \citep{Cho14} 
and a decoder with a Gaussian distribution for prediction \citep{Becker18}. 
The third baseline is a VRNN \citep{Chung15} as our base model.
The last baseline is a VRNN with macro goals as weak supervision \citep{Zhan19}, which is a state-of-the-art method in ball game trajectory prediction (but implemented as a decentralized version in Appendix~D). %  \ref{app:macro_goal}). % for a fair comparison
% The last three methods 
These are full observation models.
% without mechanical constraints.
Moreover,
% to compare the partial observation and long-term prediction performance, 
we used dynamic neural relational inference (dNRI) \citep{Graber20}, which infers a dynamic relation as a variant of NRI \citep{Kipf18} 
%,  based on each purpose in
using the basketball dataset.
We can similarly interpret our observation and the relation of dNRI. %  despite the different approaches.

%\vspace{2pt} 
%\noindent
%{\bf{Our models and tasks.}}
\subsubsection{Our models and tasks}
We validated our approach with three variants: VRNN-Mech, VRNN-Bi, and VRNN-macro-Bi-Mech. 
First, we evaluated the constraint losses and trajectory prediction performances of our mechanical constraints (denoted by ``Mech''). % using VRNN and VRNN-macro
Since our proposed binary observation model (denoted by ``Bi'') does not necessarily reduce constraint losses and improve prediction performances, we evaluated those of VRNN-Bi and VRNN-macro-Bi-Mech as a validation.
To obtain a long-term prediction, % for validation and test 
we evaluated 60 timesteps after an initial burn-in period of 20 timesteps with ground-truth states (with data sampled at 10 Hz for both datasets). 
The duration of the prediction was similar to that in \citep{Zhan19,Yeh19}. %the previous works 
We selected $x$-$y$ position, velocity, and acceleration as the input states (i.e., $d_s = 6$).
The reason for and the analysis of various inputs and outputs are described in Appendix~G. %\ref{app:dimension}.
We trained all the models using the Adam optimizer \citep{Kingma15} with default parameters using scheduled sampling to predict the next frame on the basis of on the predicted frame after the burn-in period.
% teacher forcing \citep{Williams89}.
For other training details, see Appendix~E. % \ref{app:training}.

%\vspace{2pt} 
%\noindent
%{\bf{Performance metrics.}}
\subsubsection{Performance metrics}
We evaluated several different metrics in terms of constraint losses and prediction errors to demonstrate the effectiveness of our approach on the test set.
To evaluate the mechanical plausibility, we first used the constraint losses of Eq.~(\textcolor{black}{E.1}) in Appendix~E% \ref{app:training} \ref{eq:penalty2}
, which is our specific loss function with velocity and acceleration as the output actions (validated in Appendix~G). % \ref{app:dimension}).
In this case, the penalty for the predicted dimension and the eliminated dimension can be computed only for acceleration and only for position, respectively.
The constraint losses for position and acceleration and that for smoothing acceleration are denoted as $\mathcal{L}_{pos}$,  $\mathcal{L}_{acc}$, and  $\mathcal{L}_{jrk}$, respectively.

For trajectory prediction error, we used the two basic metrics: the mean and the \textcolor{black}{smallest} $L_2$-error between the prediction and the ground truth.  
Because of the multimodal nature of the system, we randomly sampled $N =10$ trajectories for each test case.
More precisely, we computed the $L_2$-error between the ground truth and the $n$th generated sample $\hat{a}^{n,m}$ using $L_2^{n,m} = (1/T K)\sum^T
_{t=1}\sum^K_{k=1}\|\hat{a}^n_{m,t,k} - a_{m,t,k} \|_2$.
We then reported the average and best result from the samples $N$, as is standard practice (see, e.g., \cite{Rupprecht17}). % ,Bhattacharyya18
% Note that a
Although we used the ELBO for training, the tighter bounds do not necessarily lead to better performance \citep{Rainforth18}.

%\vspace{-4pt}
\subsection{Basketball data}
\label{ssec:basket}
%\vspace{-1pt}
We used the basketball dataset from the NBA 2015--2016 season (\url{ https://www.stats.com/data-science/}), which contains tracking trajectories for professional basketball players and the ball. 
The data was preprocessed such that the offense team always moved toward the left side of the court. 
We chose 100 games so that the amount of data was similar to that for the subsequent soccer dataset.
In total, the dataset contained 19968 training sequences, 2235 validation sequences, and 2608 test sequences.

\newcommand{\md}[2]{\multicolumn{#1}{c|}{#2}}
\newcommand{\me}[2]{\multicolumn{#1}{c}{#2}}

\begin{table*}[t]
\centering
%\small{
\scalebox{0.7}{
\begin{tabular}{l|rrr|rrr}%|
\Xhline{3\arrayrulewidth} %\hline
& \md{3}{Basketball data} & \me{3}{Soccer data} \\ 
& \me{1}{$\mathcal{L}_{pos}$ (NLL)} & \me{1}{$\mathcal{L}_{acc}$ (KLD)} & \md{1}{$\mathcal{L}_{jrk}$ (NLL)} & \me{1}{$\mathcal{L}_{pos}$ (NLL)} & \me{1}{$\mathcal{L}_{acc}$ (KLD)} & \me{1}{$\mathcal{L}_{jrk}$ (NLL)} \\ 
\hline
VRNN & $242.44 \pm 25.25$ & $3.70 \pm 0.22$ & $134.37 \pm 10.01$ & $564.01 \pm 50.96$ & $3.92 \pm 0.18$ & $652.42 \pm 51.39$
\\VRNN-macro &  $231.86 \pm 25.01$ & $3.67 \pm 0.21$ & $134.12 \pm 10.02$ & $363.67 \pm 27.54$ & $4.03 \pm 0.18$ & $409.12 \pm 22.65$
\\
\hline
VRNN-Mech & %
$130.81 \pm 15.46$ & $3.44 \pm 0.19$ & $124.40 \pm 8.83$ & $421.13 \pm 39.04$ & $3.84 \pm 0.17$ & $634.00  \pm 49.79$ 
%421.07 $\pm$ 38.87 & 3.84 $\pm$ 0.17 & 633.88 $\pm$ 49.81 &
\\VRNN-Bi & $273.71 \pm 28.63$ & $3.70 \pm 0.24$ & $133.75 \pm 10.25$ & $306.54 \pm 23.71$ & $3.99 \pm 0.18$ & $417.47 \pm 23.21$ 
\\VRNN-macro-Bi-Mech & $171.29 \pm 18.84$ & $3.52 \pm 0.22$ & $125.24 \pm 8.92$ & $355.26 \pm 27.26$ & $3.96 \pm 0.20$ & $404.75 \pm 22.38$
\\
\Xhline{3\arrayrulewidth} % \hline
\end{tabular}
}
\caption{\label{tab:constraint_main}NLLs and KL divergences (KLDs) as mechanical constraint losses for the basketball and soccer datasets.}
\end{table*}

%\vspace{2pt} 
%\noindent
%{\bf{Constraint losses.}}
\subsubsection{Constraint losses}
\begin{figure}[h!]% {0.7\linewidth} % sub
\centering
\includegraphics[width=0.8\columnwidth]{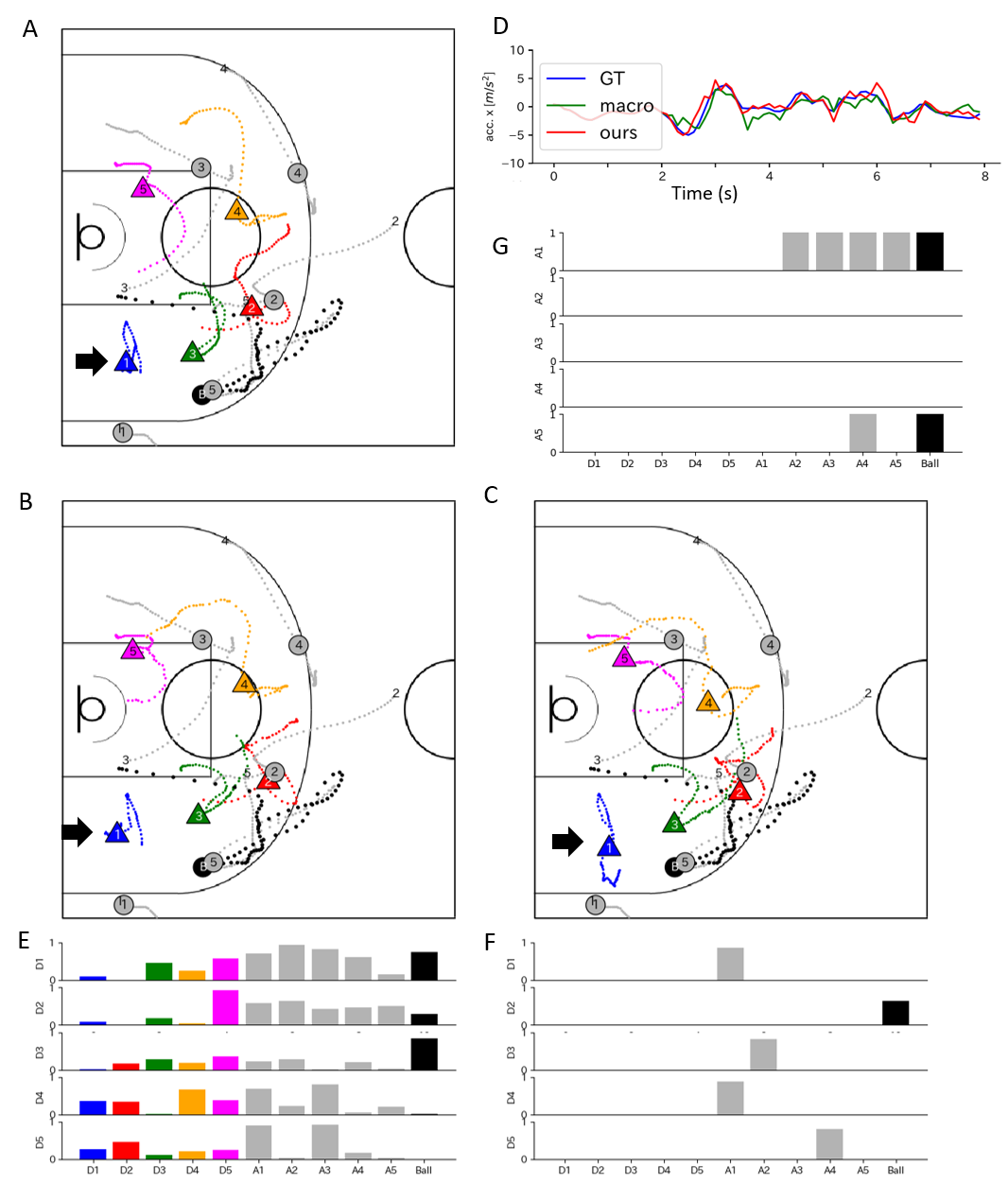}
\caption{{\small{Example results obtained with our method. (A) Ground truth, (B) a normal prediction of trajectories, and (C) a counterfactual prediction of trajectories. Colored triangles, gray circles, and the black circle are defenders, attackers, and the ball, respectively. (D) Defender~1's accelerations ($x$) for our method, the baseline, and the ground truth (GT). We evaluated 60 timesteps after an initial burn-in period of 20 timesteps with ground-truth states (at 10 Hz).  (E) and (F) Five defenders' observations $b_{t,k}$ in a normal prediction and a counterfactual prediction when they are at the marker positions on (B) and (C). (G) Relation among players and the ball obtained with dNRI with the same configuration as in (E) and (F) at the same timestamp. Since there was no relation for five defenders in this case, we show the relation for five attackers.
}}}
\label{fig:example}
\end{figure}

Table~\ref{tab:constraint_main} (left) shows our three constraint losses.
Our method (VRNN-Mech) shows smaller losses than the conventional method (VRNN). 
Although our partial observation model (VRNN-Bi) shows  losses similar to or larger than those of the baseline models (VRNN and VRNN-macro), our full model (VRNN-macro-Bi-Mech) shows smaller losses than they do.
Our model generated more physically plausible and smoother trajectories. 
For example, regarding $\mathcal{L}_{jrk}$, Fig.~\ref{fig:example}D shows examples of acceleration sequences for defender~1 (blue in Fig.~\ref{fig:example}A).
Our model (red) generated smoother accelerations like the ground truth (blue) than VRNN-macro (green).
The detailed analyses with mechanical constraints are described in Appendix~F. % \ref{app:mechanics}.

\begin{table*}[ht!]
\centering
\scalebox{0.7}{
\begin{tabular}{l|rrr|rrr}%|
\Xhline{3\arrayrulewidth} %\hline
& \md{3}{Basketball data} & \me{3}{Soccer data} \\%& \md{3}{Score prediction} \\ 
& \me{1}{Position} & \me{1}{Velocity} & \md{1}{Acceleration} & \me{1}{Position} & \me{1}{Velocity} & \me{1}{Acceleration} \\ 
\hline
Velocity & $1.41 \pm 0.34$ & $1.08 \pm 0.21$ & $10.90 \pm 2.09$ & $4.83 \pm 1.27$ & $2.72 \pm 0.46$ & $27.22 \pm 4.56$  
\\RNN-Gauss & $1.31 \pm 0.32$ & $1.05 \pm 0.13$ & $1.88 \pm 0.30$ & $2.97 \pm 0.89$ & $1.65 \pm 0.23$ & $2.22 \pm 0.33$ 
\\VRNN & $0.71 \pm 0.17$ & $0.68 \pm 0.10$ & $1.43 \pm 0.20$ &  $1.23 \pm 0.35$ & $1.06 \pm 0.21$ & $1.57 \pm 0.26$
\\VRNN-macro &  $0.71 \pm 0.17$ & $0.68 \pm 0.10$ & $1.43 \pm 0.20$ & $1.37 \pm 0.40$ & $1.11 \pm 0.21$ & $1.57 \pm 0.24$ 
\\
\hline
VRNN-Mech & $0.69 \pm 0.17$ & $0.68 \pm 0.10$ & $1.37 \pm 0.20$ & $1.22 \pm 0.34$ & $1.05 \pm 0.20$ & $1.59 \pm 0.26$
\\VRNN-Bi & $0.72 \pm 0.19$ & $0.66 \pm 0.10$ & $1.36 \pm 0.19$ & $1.20 \pm 0.37$ & $1.02 \pm 0.20$ & $1.50 \pm 0.24$ 
\\VRNN-macro-Bi-Mech & $0.73 \pm 0.18$ & $0.68 \pm 0.10$ & $1.34 \pm 0.19$ & $1.31 \pm 0.42$ & $1.06 \pm 0.21$ & $1.47 \pm 0.23$
\\
\Xhline{3\arrayrulewidth} % \hline
\end{tabular}
}
\caption{\label{tab:bestL2}\textcolor{black}{The smallest} $L_2$ prediction errors for the basketball and soccer datasets. The units for position is meters.}
\end{table*}

%\vspace{2pt} 
%\noindent
%{\bf{Trajectory prediction performances.}} 
\subsubsection{Trajectory prediction performances}
Table~\ref{tab:bestL2} (left) shows trajectory prediction performances.
We confirmed the competitive long-term prediction of our mechanical constraints (VRNN-Mech) compared with the baseline models (VRNN and VRNN-macro).
Note that our three mechanical constraints nonlinearly influence each other and do not always improve the prediction performance (it is not the main purpose).
Our observation models (VRNN-Bi and VRNN-macro-Bi-Mech) show prediction performances similar to those of the baselines regardless of the partial observation.

Additionally, we also report the trajectory prediction performance of dNRI in the same setup.
The \textcolor{black}{smallest} $L_2$ prediction errors were $2.72 \pm 0.52$, $3.02 \pm 0.34$, and $5.05 \pm 0.39$ for position, velocity, and acceleration, respectively.
These performances were worse than for Velocity or RNN-Gauss, but the purpose of \textcolor{black}{d}NRI was not long-term prediction.
The detailed analyses with mechanical constraints, discussion about VRNN-macro, and the overall mean $L_2$ prediction errors are described in Appendices~F, % \ref{app:mechanics}, 
D, % \ref{app:macro_goal}, 
and~H, %\ref{app:meanL2},
respectively.
%Note that the purpose of dNRI (and variants of NRI) is not the long-term prediction, thus the long-term prediction performances were worse than RNN in our settings.

%\noindent
%{\bf{Evaluation of observation model.}} 
\subsubsection{Evaluation of observation model}
We visually and quantitatively evaluated VRNN-macro-Bi-Mech. 
The averaged observation coefficient (i.e., binary vector $b_{t,k}$) was $4.42 \pm 0.31$ for each player (maximum 11).
That for dNRI (i.e., relation) was $1.83 \pm 0.07$.
Fig.~\ref{fig:example}E (at the moment with larger marks in Fig.~\ref{fig:example}B) shows that the defenders observed both near and far players (for the sequences for $b_{t,k}$, see Appendix~I). % \ref{app:obseq}).

Meanwhile, there was no relation for five defenders inferred by dNRI at the same timestamp (instead, we show the relation for five attackers in Fig.~\ref{fig:example}G).
Compared with our methods, dNRI extracted fewer relations or observations focusing on a specific person such as in Fig.~\ref{fig:example}G.
Quantitatively, the ratio of no relation for each defender is $(65.10 \pm 2.18)\%$, which is too large to analyze defenders' observations (that of no observation in our method is $0\%$). 
Note that since there is no ground truth of human observations or relations, accurate evaluation among methods would be difficult currently.

Our model, which can predict long-term behaviors and model observations with a large variation (near and far agents for each agent), can analyze real-world multi-agent behaviors.
For example, defender~1 (arrow) in Figs.~\ref{fig:example}A and B adopted the balanced position between attacker~1 and other attackers to help teammates near the ball (black). 
Our model can also create a counterfactual example when $b_{t,k}$ is artificially set to a one-hot vector at each moment (only the player highest in $b_{t,k}$) in Figs.~\ref{fig:example}C and~F.
In this case, like sports beginners, defender~1 went to the nearest attacker (attacker~1), and ignored the ball.
Our model can analyze such behaviors in real-world agents.
Additional analysis to compare the distances in observation among our method, dNRI, and a simple rule-based model is described in Appendix~I. % \ref{app:obseq}.

%\vspace{1pt}
\subsection{Soccer data}
%\vspace{-1pt}
To demonstrate the generality of our approach, we also used a soccer dataset \citep{Le17,Yeh19}, containing trajectories of soccer players and the ball from 45 professional soccer league games. 
We randomly split the dataset into 21504 training sequences, 2165 validation sequences, and 2452 test sequences.
We did not model the goalkeepers since they tend to move very little.
% {\bf{Prediction performance.}} 
Table~\ref{tab:constraint_main} (right) % F.2 and  %  
and Table~\ref{tab:bestL2} (right) % H.6 % Table in Appendices F and H 
indicate
% that our methods improved the constraint losses and prediction performance compared with the baselines and those of the basketball dataset. We confirmed 
the effectiveness of our mechanical constraints and the similar trajectory prediction performance of our binary observation models and the baselines, results that were similar to those for the basketball dataset.
% {\bf{Evaluation of observation.}}
For VRNN-macro-Bi-Mech, the averaged $b_{t,k}$ was $8.04 \pm 1.54$ for each defender (maximum 23).
The ratio of no relation for each defender in our method is $0\%$. 
Additional analysis to compare the distances in observation and an illustrative example are given in Appendices~I and~J%\ref{app:obseq} and  \ref{app:soccer}
, respectively. 

% ==========
\vspace*{-5pt}
\section{Conclusions}
\label{sec:conclusion} 
\vspace*{-0pt}
We proposed decentralized policy models with partial observation and mechanical constraints.
% Using real-world basketball and soccer datasets, we show the effectiveness of our method in terms of the constraint losses, long-term prediction, and partial observation.
Using real-world basketball and soccer datasets, we show the effectiveness of our method in terms of the constraint violations, long-term prediction, and partial observation.
Our approach can be used as a multi-agent simulator to generate realistic trajectories using real-world data.
One possible future research direction is to incorporate other physiological constraints into the models such as visuomotor delays and body loads, which will contribute to the understanding of smooth multi-agent behaviors in complex environments.

\section*{Acknowledgments}
This work was supported by Japan Society for the Promotion of Science KAKENHI (grant numbers 19H04941 and 23H03282) and Japan Science and Technology Agency PRESTO (grant number JPMJPR20CA).

%\section{Bibliography styles}

%There are various bibliography styles available. You can select the style of your choice in the preamble of this document. These styles are Elsevier styles based on standard styles like Harvard and Vancouver. Please use Bib\TeX\ to generate your bibliography and include DOIs whenever available.
% Here are two sample references: \cite{Feynman1963118,Dirac1953888}.

\newpage

\setcounter{page}{1}
\appendix
%\section*{Supplementary Materials}

\section{Observation in multi-agent systems.}
% Additional related works in trajectory prediction and observation models}
\label{app:observation}
In purely rule-based models, researchers investigating animals and vehicles conventionally proposed predefined observation rules based on specific distances and visual angles (e.g., \cite{Couzin02}), the specific number of the nearest agents (e.g., \cite{Ballerini08}), or other environments (e.g., \cite{Yoshihara17}). 
However, it is difficult to define interactions between agents using predefined rules in general large-scale multi-agent systems. Thus,
methods for learning observation of agents have been proposed for MARL in virtual environments \citep{Mao19,Jiang18,Iqbal19,Liu20} and for real-world multi-agent systems \citep{Hoshen17,Leurent19,Li20,Fujii20}. %Teng19

Here we introduce several approaches for neural network-based partial observation using soft and hard attention mechanisms, and other approaches.
Note that we finally aim to model a multi-hot observation in decentralized biological multi-agent systems that can also predict long-term behaviors. 

Attention is widely used in % applications including
natural language processing \citep{Bahdanau14}, computer vision \citep{Wang18}, and MARL in virtual environments \citep{Mao19,Jiang18,Iqbal19}. 
Soft attention calculates an importance distribution of elements (e.g., agents):
$w_{k}=\exp(p_k) / \sum_{i=1}^{K}\exp(p_i)$, where $p_k$ is a scalar variable for an agent $k$.
Soft attention is fully differentiable and thus can be trained with back-propagation.
However, it usually assigns nonzero probabilities to unrelated elements; i.e., it cannot directly reduce the number of agents.

Hard attention focuses solely on an important element but is basically nondifferentiable because of the selection based on sampling.
Among differentiable models with discrete variables, an approach using Gumbel-softmax reparameterization \citep{Jang17,Maddison17} can be effectively implemented via a continuous relaxation of a discrete categorical distribution. 
In summary, given $K$-categorical distribution parameters $p$,
a differentiable $K$-dimensional one-hot encoding sample $G$ from the Gumbel-softmax distribution can be computed as:
%\begin{align*}
$G(\text{log} p)_k = \text{exp}((\text{log} p_k + \eps)/\tau ) / \sum_{i=0}^K \text{exp}((\text{log} p_i + \eps)/\tau ) ,$ % \frac{
%\end{align*}
where $\eps$ are independent identically
distributed samples from a Gumbel$(0,1)$ distribution, i.e., 
%\begin{align*}
$\eps = -\text{log}(-\text{log}(u)), \; \; u \sim \mathcal{U}[0,1]$.
%\end{align*}
$\mathcal{U}$ is a uniform distribution and $\tau$ is the softmax temperature parameter. 
However, the resulting one-hot vector $[G(\text{log} p)_1,\ldots,G(\text{log} p)_K]$ is insufficient to represent a multi-hot observation.

%%%%%%%%%%%%%%%%%%%%%%%%%%%%
\section{Role assignment problem}
\label{app:roleassign}
\vspace{-0pt}

In most real-world data, training (i.e., demonstration) data and test data include sequences from different types of agent (e.g., teams and games). 
Learning policies based on players' positions or roles, instead of identity, seems to be more natural and more data-efficient in general.
Among several approaches \citep{Le17,Yeh19}, for the behavioral modeling in a POMDP, we separate the problem into role assignment problem and policy learning based on \citep{Le17}.

The unstructured set of demonstrations is denoted by $u=\{u_{1},\ldots, u_{K}\}$, where $u_k = \{ u_{t,k}\}_{t=1}^T$ is the sequence of actions by agent $k$ at time $t$.  
Let $c = \{ c_t\}_{t=1}^T$ be the context associated with each demonstration sequence (e.g., an opponent team and ball). 
In a role assignment problem, the indexing mechanism is formulated as an assignment function $\mathcal{A}$ that maps the unstructured set $u$ and some probabilistic structured model $q$ to an indexed set of states $s$ rearranged from $u$; i.e.,
\begin{equation*}
\mathcal{A}:\{ u_1,\ldots,u_K, c\}\times q\mapsto \left[ s_1,\ldots,s_K,c\right],
\end{equation*}
where the set $\{ s_1,\ldots,s_K\} \equiv \{ u_1,\ldots,u_K\}$.
We consider $q$ as a latent variable model that infers the role assignments for each set of demonstrations.
The role assignment with a latent structured model addresses two main issues: (1) unsupervised learning of a probabilistic role assignment model $q$; and (2) the indexing with $q$ so that unstructured sequences can be mapped to structured sequences.

First, in the unsupervised learning of the stochastic role assignment model, we use a Gaussian hidden markov model (Gaussian HMM) according to \citep{Le17}.
The Gaussian HMM inputs a state feature vector defined by all agents' information and learn transition probabilities and Gaussian mixture distributions as output probabilities of hidden states.

Second, for indexing based on $q$ to map $u$ to $s$,
we solve a well-known linear assignment problem \citep{Papadimitriou82}.
Concretely, the distance between the mixed Gaussian distribution obtained above and the state feature vector is computed for each time and each player.
The linear assignment problem is then solved with use of the distance as a cost function (i.e., roles are assigned in order of players whose distance is closer to each Gaussian distribution).

%%%%%%%%%%%%%%%%%%%%%%%%%%%%%%%%%%%%%%
\vspace{-0pt}
\section{Variational recurrent neural networks}
\label{app:sequential}
\vspace{-0pt}

Here we briefly overview RNNs, VAEs, and VRNNs.

% \paragraph{.}
%
From the perspective of a probabilistic generative model, an RNN models the conditional probabilities with a hidden state $h_t$ that summarizes the history in the first $t-1$ timesteps:
\eq{
p_{\theta}(a_t | a_{< t}) = \varphi(h_{t-1}), \quad \quad h_t = f(a_t, h_{t-1}),
}
where $\varphi$ maps the hidden state to a probability distribution over states and $f$ is a deterministic function such as LSTMs %\citep{Hochreiter97} 
or gated recurrent units (GRUs). %\citep{Cho14}. 
RNNs with simple output distributions often struggle to capture highly variable and structured sequential data. Recent work in sequential generative models addressed this issue by injecting stochastic latent variables into the model and using amortized variational inference to infer latent variables from data.
A VRNN \citep{Chung15} is one of the methods using this idea and combining RNNs and VAEs. 

A VAE \citep{Kingma14} is a generative model for nonsequential data that injects latent variables $z$ into the joint distribution $p_{\theta}(a, z)$ and introduces an inference network parameterized by $\phi$ to approximate the posterior $q_{\phi}(z \mid a)$. 
The learning objective is to maximize the ELBO of the log-likelihood with respect to the model parameters $\theta$ and $\phi$:
\eq{\mathbb{E}_{q_{\phi}(z | a)}\brcksq{\log p_{\theta}(a | z)} - D_{KL}(q_{\phi}(z \mid a) || p_{\theta}(z))}.
The first term is known as the reconstruction term and can be approximated with Monte Carlo sampling. 
The second term is the KL divergence between the approximate posterior and the prior, and can be evaluated analytically if both distributions are Gaussian with diagonal covariance. 
The inference model $q_{\phi}(z \mid a)$, generative model $p_{\theta}(a \mid z)$, and prior $p_{\theta}(z)$ are often implemented with neural networks.

VRNNs combine VAEs and RNNs by conditioning the VAE on a hidden state $h_t$:
\eq{
p_{\theta}(z_t | a_{<t}, z_{<t}) & = \varphi_{\text{prior}}(h_{t-1}) & & \text{(prior)} \label{eq:vrnn_prior} \\
q_{\phi}(z_t | a_{\leq t}, z_{<t}) & = \varphi_{\text{enc}}(a_t, h_{t-1}) & & \text{(inference)} \\
p_{\theta}(a_t | z_{\leq t}, a_{<t}) & = \varphi_{\text{dec}}(z_t, h_{t-1}) & & \text{(generation)} \\
h_t & = f(a_t, z_t, h_{t-1}). & & \text{(recurrence)}
\label{eq:vrnn_state}
}
VRNNs are also trained by maximizing the ELBO, which can be interpreted as the sum of VAE ELBOs over each timestep of the sequence:
\eq{
\mathbb{E}_{q_{\phi}(z_{\leq T} \mid a_{\leq T})} & \Bigg[ \sum_{t=1}^T \log p_{\theta}(a_t \mid z_{\leq T}, a_{<t}) \\
& - D_{KL} \Big( q_{\phi}(z_t \mid a_{\leq T}, z_{<t}) || p_{\theta}(z_t \mid a_{<t}, z_{<t}) \Big) \Bigg] \nonumber
% \label{eq:vrnn_elbo_appendix}
}
Note that the prior distribution of latent variable $z_t$ depends on the history of states and latent variables (Eq. (\ref{eq:vrnn_prior})).

%%%%%%%%%%%%%%%%%%%%%%%%%%%%%%%%%%%%%%
\vspace{-0pt}
\section{Decentralized and partially observed macro goals}
\label{app:macro_goal}
\vspace{-0pt}

Our model uses macro goal \citep{Zhan19} for 
long-term prediction by modifying our decentralized and partially observable setting.
Here we briefly review the original macro goal and describe our modification of the decentralized and partially observable setting. We also discuss the related results of our experiments.

As an illustrative example, Fig. 2B shows macro goals for a basketball defender as specific areas on the court (boxes). 
After reaching the macro goal in the center, the blue player moves towards the next macro goal at the top (middle).
The macro goals provide a compact summary of the players' sequences over a long time to encode long-term intent.

\subsection{Shared macro goals}
The original macro goal or macro intent \citep{Zheng16,Zhan19}, obtained via some labeling functions, is defined as low-dimensional and spatiotemporal representations of the data for the learning of multi-agent long-term coordination such as basketball (Figure 2B).
The original macro goal assumes that: (1) it provides a tractable way to capture coordination between agents; (2) it encodes long-term goals of agents and enables long-term planning at a higher-level timescale; and (3) it compactly represents some low-dimensional structure in an exponentially large multi-agent state space.

Specifically, the modeling assumptions for the original macro goals are as follows: 1) agent states $s_{t}$ in a period $[t_1, t_2]$ are conditioned on some shared macro goal $g_t$; 
2) the start and end times $[t_1, t_2]$ of episodes can vary between sequences; 
3) macro goals change slowly over time relative to the agent states: $d g_t / dt \ll 1$, and 
4) due to their reduced dimensionality, (near-) arbitrary dependencies between macro goals (e.g., coordination) can be modeled by a neural network approach.

Among various labeling approaches (see \citep{Zhan19}) for the macro goal, programmatic weak supervision is a method requiring low labor cost, effectively learning the underlying structure of large unlabeled datasets, and allowing users to incorporate domain knowledge into the model. 
For example, the labeling function to obtain macro goals for basketball sequences computes the regions on the court in which players remain stationary; this integrates the idea that players aim to set up specific formations on the court.

Specifically, previous work labeled the macro goal independently among agents (obtained in a rule based manner), and learned the shared macro goal model via supervised learning by maximizing the log-likelihood of macro goal labels.
The model finally returns the one-hot encoding of the box that contains the position information.

%
% \textbf{Decentralized and partially observed macro goals.}
\subsection{Specific setup in our experiments}
Our decentralized and partially observed macro goals (1) use partial observation $o_{t-1,k}$ to obtain the macro goal model and 
(2) independently learn the decentralized macro goal $g'_{t,k}$ (not shared between agents).
Among several labeling functions, we adopted the stationary labeling function to compute the macro goal on the basis of stationary positions because it reflects important information about the structure of the data, and its better performance was confirmed \citep{Zhan19}.

For basketball data, according to the previous work \citep{Zhan19}, we define the macro goal by segmenting the left half-court into a $10 \times 9$ grid of $5\,\mathrm{feet} \times 5\,\mathrm{feet}$ boxes (Fig.~2B). 
For soccer data, we segmented the total court into a $34 \times 22$ grid of approximately $3\,\mathrm{m} \times 3\,\mathrm{m}$ boxes. 

\subsection{Related discussion in our experiments}
We confirmed the decentralized and partially observed macro goals did not improve the prediction performances.
There are mainly two reasons for this.
One is obviously the decentralized (i.e., not shared) setting, but it is a necessary assumption for our modeling.
The second is the improvement of the VRNN baseline by our adding dropout and batch normalization layers to avoid overfitting (note that they were added in all models for fair comparisons).
Specifically in the soccer experiment, the resolution of the grid might make it larger than that of the prediction (but the smaller grid might be difficult to learn the macro goal).

%%%%%%%%%%%%%%%%%%%%%%%%%%%%%%%%%%%%%%

%%%%%%%%%%%%%%%%%%%%%%%%%%%%%%%%%%%%%%
\vspace{-0pt}
\section{Training details}
\label{app:training}
\vspace{-0pt}
Here we describe a specific objective function in our experiments, and other training details.
For other implementation details, such as preprocessing including role assignment in \ref{app:roleassign}, see the code at \url{https://github.com/keisuke198619/PO-MC-DHVRNN}.

\subsection{Objective function in our experiments}
We designed the objective function as described in Section 3.4, % \ref{ssec:learning}, 
but we specially weighted each term of the objective function.
Note that, in our experiments, we selected velocity and acceleration as the output actions (see \ref{app:dimension}).
In this case, the penalty for the predicted dimension and the eliminated dimension can be computed only for acceleration and only for position, respectively.
That is, the weighted penalties of the mechanical constraints are 
\vspace{-2pt}
\begin{align}
&\mathcal{L}_{w\mathchar`-body} =   \mathbb{E}_{\theta} \sum_{t=2}^T \Big[ \label{eq:penalty2}\\
&\lambda_{acc} D_{KL} ( p_{\theta}(\hat{a}_{acc,t} \mid z_{\leq t}, o_{<t}, g^{\prime}_{\leq t}) || p_{\theta}(\tilde{a}_{acc,t} \mid z_{\leq t}, o_{<t}, g^{\prime}_{\leq t}) ) \notag\\
& - \lambda_{pos} \log p_{\theta}(a_{pos,t} \mid z_{\leq t}, o_{<t}, g^{\prime}_{<t})
\notag\\
& - \lambda_{jrk} \log p_{\theta}(a_{acc,t+1} \mid z_{\leq t}, o_{<t}, g^{\prime}_{<t}) \Big],
\notag
\vspace{0pt}
\end{align}
where $\lambda_{acc}$, $\lambda_{pos}$, and $\lambda_{jrk}$ are regularization parameters.
As shown in \ref{app:mechanics}, the first and second terms improved the prediction performance of velocity and position, respectively (the third term basically contributed to smoothing in acceleration as shown in Fig. 3D). %  \ref{fig:example}
The results also show that the learning of acceleration was relatively difficult because of the above-mentioned imbalance between the dimensions.
We then added the reconstruction term $\mathcal{L}_{rec}$ in $\mathcal{L}_{vrnn}$ only for acceleration. 
In summary, the specific objective function in our experiments was $\mathcal{L}_{vrnn} + \mathcal{L}_{w\mathchar`-body} + \lambda_{rec}\mathcal{L}_{rec} $.
We set $\lambda_{acc} = 0.1$, $\lambda_{pos} = 0.01$, $\lambda_{jrk} = 0.1$, and $\lambda_{rec} = 0.2$ in the basketball experiment and $\lambda_{acc} = 0.01$, $\lambda_{pos} = 0.01$, $\lambda_{jrk} = 0.02$, and $\lambda_{rec} = 0.001$ in the soccer experiment, because the trajectories in the soccer dataset were more difficult to predict than those in the basketball dataset as shown in Table 2 (i.e., the reconstruction was prioritized).

\vspace{-0pt}
\subsection{Other training details}
We trained all the models using the Adam optimizer \citep{Kingma15} with default parameters using teacher forcing \citep{Williams89}.
To prevent overfitting, dropout and batch normalization layers were used (the dropout rate was set to 0.5), and we selected the model with the best performance on the validation set.
% All fully connected layers are initialized using Xavier initialization \citep{Glorot10}.
We set the embedding dimension to $d_e = 32$ for each agent (and the ball).
The embedding layer, the macro goal decoder $\varphi_{g'_k}$, prior $\varphi^k_{\text{prior}}$, encoder $\varphi^k_{\text{enc}}$, and decoder $\varphi^k_{\text{dec}}$ in the VRNN were implemented by a two-layer neural network with a hidden layer of size $64$.
We modeled each latent variable $z$ as a multivariate Gaussian with diagonal covariance of dimension $64$. 
All GRUs were implemented by a two-layer neural network with a hidden layer of size $100$.
Other implementations were based on \citep{Zhan19}.
We selected $x-y$ position, velocity, and acceleration as the input states (i.e., $d_s = 6$).
% We selected velocity and acceleration as the output actions.
The reason for and the analysis of various input states and output actions are described in \ref{app:dimension}.
For the temperature $\tau$ in a Gumbel-softmax distribution, we set it to $1$ in both experiments.
% the basketball experiment and $0.1$ in the soccer experiment.
%This is because the latter includes more unnecessary information for the prediction (in general, the lower temperature is set, the more one-hot vector is obtained).

The VRNN decoder $\varphi^k_{\text{dec}}$ was implemented by a two-layer neural network returning a multivariate Gaussian with diagonal covariance. 
The original VRNN \citep{Chung15} has no constraint in the learning of the variance.
It may cause the NLL $\log p_{\theta}(a_t \mid z_{\leq t})$ to tend toward infinity when the variance approaches zero. 
Most of the cases, including our experiments, such problems did not happen (in such difficult cases, setting the decoder variance as a global hyperparameter will be practically effective).
In our model, the KL divergence in the first term in Eq. (3) includes $-2\log (\hat{\sigma}^{t,k}_{dec})$, where $\hat{\sigma}^{t,k}_{dec}$ is the directly estimated variance, and the term may prevent the variance from approaching zero. % \ref{eq:penalty}

\subsection{Training in dNRI}
Additionally, we also investigated dNRI \citep{Graber20} as a baseline, which infers dynamic relation as a variant of NRI \citep{Kipf18}, based on the same setting (2-s burn-in and 6-s prediction of five defenders) using the basketball dataset.
The batch size was 32, and the other hyperparameters were the same as in previous work \citep{Graber20}.
Note that our methods modeled five defenders independently, but dNRI modeled all agents and the ball (i.e., regarding five attackers and the ball, the ground-truth data is used also in the test prediction phase).
Therefore, dNRI can infer the attackers' (and the ball's) relations, such as in Fig. 3G. %\ref{fig:example}G.

%%%%%%%%%%%%%%%%%%%%%%%%%%%%%%%%%%%%%%
\vspace{-0pt}
\section{Analysis of mechanical constraints}
\label{app:mechanics}
\vspace{-0pt}

Here we verify various mechanical constraints in our model.
The detailed objective function was described in \ref{app:training}.
For clarity, we evaluated the prediction performance of the VRNN using the basketball data among various options: (1) VRNN, (2) VRNN-$C_{pos}$, (3) VRNN-$C_{pos,acc}$, (4) VRNN-$C_{pos,acc,jrk}$, and (5) VRNN-$C_{pos,acc,jrk}$-$\mathcal{L}_{rec}$ (VRNN-Mech in the main text). 
The second, third, and fourth options added the penalty of the second, first, and third terms in Eq. (\ref{eq:penalty2}), respectively.
The fifth option added the acceleration reconstruction term $\mathcal{L}_{rec}$.

Table \ref{tab:constraint_loss} gives the results for the constraint losses for the basketball dataset. 
Obviously, $\mathcal{L}_{pos}$, $\mathcal{L}_{acc}$, and $\mathcal{L}_{jrk}$ improved by addition of $C_{pos}$, $C_{acc}$, and $C_{jrk}$, respectively. 
When $\mathcal{L}_{rec}$ was added, $\mathcal{L}_{acc}$ (and the prediction of acceleration below) improved but $\mathcal{L}_{jrk}$ became worse. 
In other words, there was a trade-off between $\mathcal{L}_{jrk}$ and the prediction of acceleration.

\begin{table}[h]
\centering
%\small{
\scalebox{0.75}{
\begin{tabular}{l|rrr|rrr}%|
\Xhline{3\arrayrulewidth} %\hline
% & \me{3}{Basketball data} \\% & \me{3}{Soccer data} \\%& 
& \me{1}{$\mathcal{L}_{pos}$(NLL)} & \me{1}{$\mathcal{L}_{acc}$(KLD)} & \me{1}{$\mathcal{L}_{jrk}$(NLL)}   \\ 
\hline
VRNN & 242.44 $\pm$ 25.25 & 3.70 $\pm$ 0.22 & 134.37 $\pm$ 10.01 
\\VRNN-$C_{pos}$ &\textbf{124.83 $\pm$ 15.87} & 4.01 $\pm$ 0.17 & 115.78 $\pm$ 9.62
\\VRNN-$C_{pos,acc}$ & 162.29 $\pm$ 18.60 & 3.69 $\pm$ 0.17 & 111.03 $\pm$ 8.53 
\\VRNN-$C_{pos,acc,jrk}$ & 156.05 $\pm$ 18.28 & 3.69 $\pm$ 0.17 & \textbf{107.22 $\pm$ 8.13} 
\\
\hline
VRNN-$C_{pos,acc,jrk}$-$\mathcal{L}_{rec}$ & %
130.81 $\pm$ 15.46 & \textbf{3.44 $\pm$ 0.19} & 124.40 $\pm$ 8.83 
\\
\Xhline{3\arrayrulewidth} % \hline
\end{tabular}
}
\caption{\label{tab:constraint_loss} {\small{NLLs and KL divergences (KLDs) as mechanical constraint losses among various constraints.}}}
\end{table}

Table~\ref{tab:constraint} gives the results for prediction performances.
$\mathcal{L}_{acc}$ slightly improved the prediction performance of velocity ($\mathcal{L}_{jrk}$ basically contributed to smoothing in acceleration as shown in Fig.~3D). % \ref{fig:example}D).
% VRNN-$C_{pos,acc,jrk}$ (VRNN-Mech in the main text) shows a well-balanced prediction performance in all dimensions.
The results also show that the learning of acceleration was relatively difficult because of the above-mentioned imbalance between the dimensions.
We thus added the reconstruction term $\mathcal{L}_{acc}$. 
VRNN-$C_{pos,acc,jrk}$-$\mathcal{L}_{acc}$ (VRNN-Mech in the main text) shows better prediction performance in all dimensions.
 
\begin{table*}[ht!]
\centering
% \small{
\scalebox{0.7}{
\begin{tabular}{l|rrr|rrr}%|
\Xhline{3\arrayrulewidth} %\hline
& \md{3}{Average $L_2$} & \me{3}{\textcolor{black}{Smallest} $L_2$} \\%& \md{3}{Score prediction} \\ 
& \me{1}{Position} & \me{1}{Velocity} & \md{1}{Acceleration} & \me{1}{Position} & \me{1}{Velocity} & \me{1}{Acceleration} \\ 
\hline
VRNN & $0.90 \pm 0.18$ & $0.73 \pm 0.10$ & $1.52 \pm 0.20$ & $0.71 \pm 0.17$ & $0.68 \pm 0.10$ & $1.43 \pm 0.20$
\\VRNN-$C_{pos}$ & $0.90 \pm 0.18$ & $0.74 \pm 0.10$ & $1.49 \pm 0.20$ & $0.72 \pm 0.17$ & $0.69 \pm 0.10$ & $1.40 \pm 0.20$
\\VRNN-$C_{pos,acc}$ & $0.88 \pm  0.18$ & $0.71 \pm  0.10$ & $1.54 \pm  0.21$ & $0.70 \pm 0.17$ & $0.66 \pm 0.10$ & $1.45 \pm 0.21$ 
\\VRNN-$C_{pos,acc,jrk}$ & $0.89 \pm 0.18$ & $0.71 \pm 0.10$ & $1.54 \pm 0.21$ & $0.71 \pm 0.17$ & $0.66 \pm 0.10$ & $1.45 \pm 0.21$
\\
\hline
VRNN-$C_{pos,acc,jrk}$-$\mathcal{L}_{acc}$ % (Mech)
& $0.87 \pm 0.18$ & $0.72 \pm$ 0.10 & $1.46 \pm$ 0.21 & $0.69 \pm 0.17$ & $0.68 \pm 0.10$ & $1.37 \pm 0.20$
% VRNN-$C_{pos,acc,jrk}$ (Mech) & $0.73 \pm 0.22$ & $0.69 \pm 0.13$ & $1.19 \pm 0.23$ & $0.57 \pm 0.19$ & $0.66 \pm 0.12$ & $1.12 \pm 0.22$  
\\ % \hline
% VRNN-$C_{pos,acc,jrk}$-$\mathcal{L}_{acc}$ (Mech) & $0.90 \pm 0.21$ & $0.99 \pm 0.18$ & $1.08 \pm 0.20$ & $0.75 \pm 0.20$ & $0.95 \pm 0.18$ & $1.00 \pm 0.19$ \\
\Xhline{3\arrayrulewidth} % \hline
\end{tabular}
}
\caption{\label{tab:constraint} {\small{Average and \textcolor{black}{smallest} $L_2$ prediction errors for various constraints obtained with the basketball dataset.}}}
\vspace{-0pt}
\end{table*}

%%%%%%%%%%%%%%%%%%%%%%%%%%%%%%%%%%%%%%
\vspace{-0pt}
\section{Analysis of input state and output action dimensions}
\label{app:dimension}
\vspace{-0pt}

Here we verify various options to select input state and output action dimensions among position, velocity, and acceleration.
The action in previous studies (e.g., \cite{Zhan19,Yeh19}) was usually the agent's position, but these studies reported the difficulty in learning velocity and acceleration (e.g., change in direction of movement), which is critical in our problem. 
We aim to obtain policy models to generate biologically realistic actions in terms of position, velocity, and acceleration.

For simplicity, we evaluated the prediction performance of the VRNN using the basketball data among various options: (1) VRNN-pos, (2) VRNN-vel, (3) VRNN-acc, (4) VRNN-pos-vel-acc, and (5) VRNN-vel-acc.
The first one uses only positional information as the input and output as the baseline as used in most of the previous studies (e.g., \cite{Zhan19,Yeh19}).
The second one uses position and velocity as the input and only velocity as the output for comparison with VRNN-pos and VRNN-vel-acc. 
The remaining ones use the full information (position, velocity, and acceleration) as the input.
The third and fourth ones use only acceleration and the full information as the output, respectively, for comparison with VRNN-vel-acc.
The fifth one uses velocity and acceleration as the output.
The second, third, and fifth ones compute the future positions using the current position and velocity (the third similarly computes velocity).

Table~\ref{tab:dimension} gives the results of the analysis.
The model effectively learned the dimension of the output (e.g., position in VRNN-pos) but the indirect learning of the differential value (e.g., velocity) was difficult.
In contrast, for the indirect learning of the integral value, the learning of position using velocity was effective, but the learning of velocity using acceleration was difficult.
This may be caused by the data having noisy accelerations.
The specific difficulty in numerical differentiation and integration, which is considered as a hard constraint in Section~3.3, %\ref{ssec:policy},
derived possibly from the low spatiotemporal resolution of the data. 
The learning of all dimensions (VRNN-pos-vel-acc) did not show better performance in all dimensions.
VRNN-vel-acc shows better prediction performance in all dimensions.
Therefore, we chose VRNN-vel-acc as our base model.

\begin{table*}[ht!]
\centering
% \small{
\scalebox{0.7}{
\begin{tabular}{l|rrr|rrr}%|
\Xhline{3\arrayrulewidth} %\hline
& \md{3}{Average $L_2$} & \me{3}{\textcolor{black}{Smallest} $L_2$} \\%& \md{3}{Score prediction} \\ 
& \me{1}{Position} & \me{1}{Velocity} & \md{1}{Acceleration} & \me{1}{Position} & \me{1}{Velocity} & \me{1}{Acceleration} \\ 
\hline
VRNN-pos & $1.60 \pm 0.33$ & $16.01 \pm 3.26$ & $160.12 \pm 32.59$ & $1.50 \pm 0.31$ & $15.05 \pm 3.10$ & $150.46 \pm 31.01$ 
\\VRNN-vel & $0.75 \pm 0.17$ & $0.58 \pm 0.09$ & $5.76 \pm 0.88$ & $0.63 \pm 0.16$ & $0.54 \pm 0.09$ & $5.40 \pm 0.85$
\\VRNN-acc & $1.61 \pm 0.34$ & $0.80 \pm 0.13$ & $1.21 \pm 0.17$ & $1.23 \pm 0.31$ & $0.68 \pm 0.12$ & $1.13 \pm 0.17$ 
\\VRNN-pos-vel-acc & $0.90 \pm$ 0.18 & $0.73 \pm 0.10$ & $1.52 \pm 0.20$ & $0.71 \pm 0.17$ & $0.68 \pm 0.10$ & $1.43 \pm 0.20$  
\\
\hline
VRNN-vel-acc (main) & $1.08 \pm 0.23$ & $0.75 \pm 0.11$ & $1.32 \pm 0.19$ & $0.85 \pm 0.21$ & $0.70 \pm 0.11$ & $1.23 \pm 0.19$ 
\\
\Xhline{3\arrayrulewidth} % \hline
\end{tabular}
}
\caption{\label{tab:dimension} {\small{Average and \textcolor{black}{smallest} $L_2$ prediction errors for various inputs and outputs obtained with the basketball dataset.}}}
\vspace{-0pt}
\end{table*}

%%%%%%%%%%%%%%%%%%%%%%%%%%%%%%%%%%%%%%
\vspace{-0pt}
\section{Supplemental results of prediction error}
\label{app:meanL2}
\vspace{-0pt}

The average $L_2$ prediction errors in Table~\ref{tab:meanL2} indicate similar tendencies as the \textcolor{black}{smallest} $L_2$ prediction errors in Table~2. % \ref{tab:bestL2}.

Additionally, we also examined dNRI \citep{Graber20}.
The average $L_2$ prediction errors were $4.57 \pm 0.44$, $6.42 \pm 0.35,$ and $8.34 \pm 0.50$ in position, velocity, and acceleration, respectively.
These performances were worse than for Velocity or RNN-Gauss, but the purpose of dNRI (and variants of NRI) was not long-term prediction.

\begin{table*}[ht!]
\centering
% \small{
\scalebox{0.7}{
\begin{tabular}{l|rrr|rrr}%|
\Xhline{3\arrayrulewidth} %\hline
& \md{3}{Basketball data} & \me{3}{Soccer data} \\%& \md{3}{Score prediction} \\ 
& \me{1}{Position} & \me{1}{Velocity} & \md{1}{Acceleration} & \me{1}{Position} & \me{1}{Velocity} & \me{1}{Acceleration} \\ 
\hline
Velocity & $1.41 \pm 0.34$ & $1.08 \pm 0.21$ & $10.90 \pm 2.09$ & $4.83 \pm 1.27$ & $2.72 \pm 0.46$ & $27.22 \pm 4.56$  
\\RNN-Gauss & $1.57 \pm 0.34$ & $1.12 \pm 0.14$ & $1.95 \pm 0.31$ & $3.19 \pm 0.90$ & $1.71 \pm 0.23$ & $2.28 \pm 0.34$ 
\\VRNN & $0.90 \pm 0.18$ & $0.73 \pm 0.10$ & $1.52 \pm 0.20$  & $1.46 \pm 0.39$ & $1.11 \pm 0.21$ & $1.63 \pm 0.27$
% \\VRNN-macro (pos) &
\\VRNN-macro &  $0.90 \pm 0.18$ & $0.73 \pm 0.10$ & $1.53 \pm 0.21$ & $1.63 \pm 0.44$ & $1.17 \pm 0.21$ & $1.63 \pm 0.25$
\\
\hline
VRNN-Mech & $0.87 \pm 0.18$ & $0.72 \pm 0.10$ & $1.46 \pm 0.21$ & $1.44 \pm 0.38$ & $1.10 \pm 0.21$ & $1.65 \pm 0.27$
\\VRNN-Bi & $0.90 \pm 0.20$ & $0.71 \pm 0.11$ & $1.44 \pm 0.19$ & $1.41 \pm 0.41$ & $1.07 \pm 0.20$ & $1.55 \pm 0.24$
% \\VRNN-macro-Mech & 
\\VRNN-macro-Bi-Mech & $0.93 \pm 0.20$ & $0.73 \pm 0.11$ & $1.42 \pm 0.19$ & $1.55 \pm 0.46$ & $1.11 \pm 0.22$ & $1.53 \pm 0.24$ 
\\
\Xhline{3\arrayrulewidth} % \hline
\end{tabular}
}
\caption{\label{tab:meanL2} Average $L_2$ prediction errors for basketball and soccer datasets.}
\vspace{0pt}
\end{table*}

%%%%%%%%%%%%%%%%%%%%%%%%%%%%%%%%%%%%%%
\section{Additional analysis of observations}
\label{app:obseq}
To compare our method with a simple rule-based model, we computed three distances between the subject player and the furthest player observed in our model, dNRI, and a rule-based model.
The last considered the same-order nearest player as the number of players observed in our model and dNRI (different numbers), which is considered to be determined by a compatible and well-known predetermined rule \citep{Ballerini08}.
The mean distance from the furthest player in our model was greater than that of the same-order nearest player ($7.56\,\mathrm{m} \pm 1.30\,\mathrm{m}$ and $3.39\,\mathrm{m} \pm 0.81\,\mathrm{m}$).
In dNRI, it was relatively greater than that of the same-order nearest player ($6.84\,\mathrm{m} \pm 1.99\,\mathrm{m}$ and $2.10\,\mathrm{m} \pm 1.05\,\mathrm{m}$).
Our model and dNRI reflected far agent information (but we ignored no-relation sequences in dNRI).
For example, defender~1 (arrow) in Figs.~3A and~B adopted the balanced position between attacker~1 and other attackers to help teammates near the ball (black). % \ref{fig:example}

In the soccer dataset, our model reflected far agent information (the mean distance in our model was $26.59\,\mathrm{m} \pm 5.44\,\mathrm{m}$ and that in the same-order nearest player was $18.09\,\mathrm{m} \pm 6.52\,\mathrm{m}$).

\begin{figure*}[h]%{0.48\linewidth} % sub
\centering
\includegraphics[width=1\columnwidth]{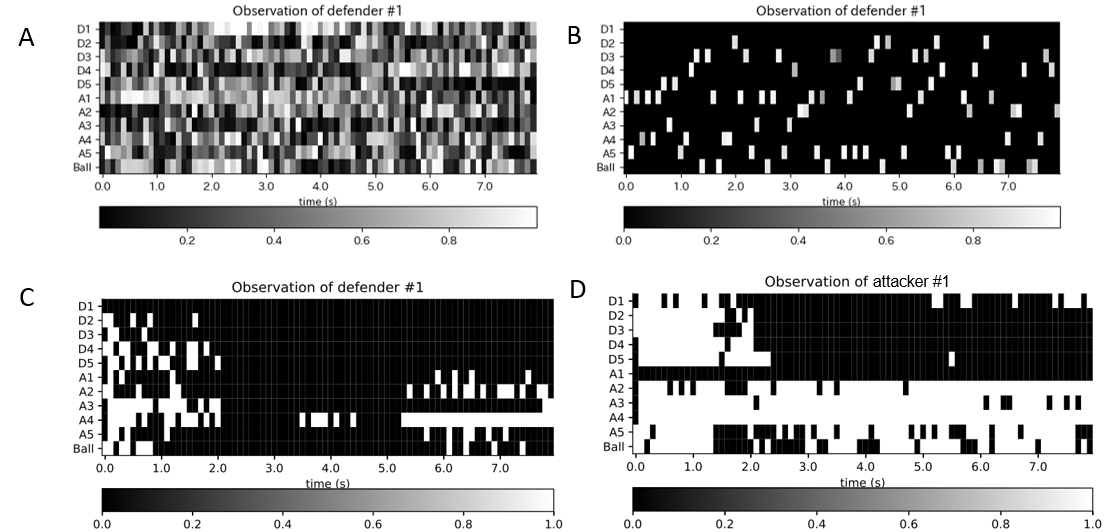}
\caption{{\small{Example sequences of the observation coefficients for defender~1 in (A) our normal prediction, (B) our counterfactual prediction, and (D) the dNRI prediction. These correspond to Figs.~3E, F, and~G, respectively. % \ref{fig:example}
(C) Defender~1's relation inferred by dNRI but at the moment in Fig.~3 ($t=0.34\,\mathrm{s}$), there was no observation.}}} %  \ref{fig:example}
\label{fig:obseq}
\vspace{-12pt}
\end{figure*}

Next, we show example sequences of the observation coefficients in Figs.~\ref{fig:obseq}A, B, and~D, which correspond to defender~1's observation coefficients in Figs.~3E, F, and~G (our normal prediction, our counterfactual prediction, and the dNRI prediction for the basketball dataset), respectively. % \ref{fig:example}
Fig.~\ref{fig:obseq}C shows the defender~1's relation inferred by dNRI, but at the moment in Fig.~3 ($t=0.34\,\mathrm{s}$), there was no observation. % \ref{fig:example}
We confirmed that the observation coefficients for the ball and the nearest attacker (attacker~1) were higher than for other players in Figs.~\ref{fig:obseq}A and~B.
The defender's relation extracted by dNRI in Fig.~\ref{fig:obseq}C was sparser than for our observation model in Fig.~\ref{fig:obseq}A, whereas the attacker's relation in Fig.~\ref{fig:obseq}D was dense. 
However, quantitatively, the ratio of no relation for each attacker in dNRI is $(63.54 \pm 2.62)\%$, which is a level similar to that for each defender described in Section~4.2. % \ref{ssec:basket}. 

%%%%%%%%%%%%%%%%%%%%%%%%%%%%%%%%%%%%%%
\vspace{-0pt}
\section{Evaluation results and an illustrative example using the soccer dataset}
\label{app:soccer}
\vspace{-0pt}

\begin{figure*}[h]%{0.48\linewidth} % sub
\centering
\includegraphics[width=1\columnwidth]{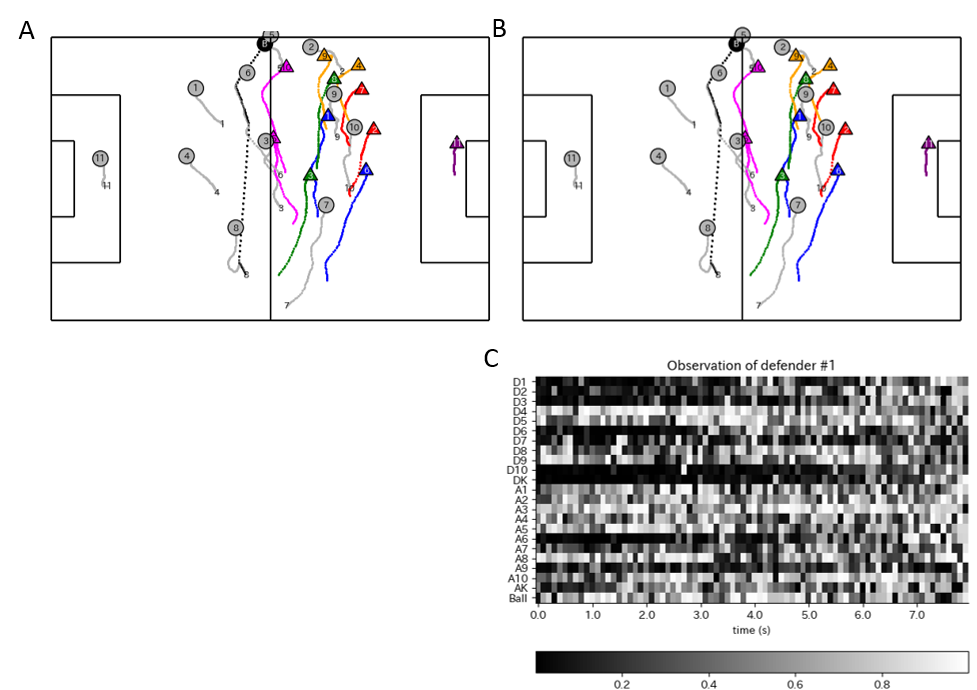}
\caption{{\small{Example results obtained with our method. (A) Ground truth and (B) a normal prediction. Colored triangles, gray circles, and the black circle are defenders, attackers, and the ball. (C) Defender~1's observations in a normal prediction.}}}
\label{fig:soccer}
\vspace{-12pt}
\end{figure*}

We obtained an example for VRNN-macro-Bi-Mech using the soccer dataset in Fig.~\ref{fig:soccer}.
Unlike the basketball example in Fig.~3, % \ref{fig:example}, 
the ground truth and a normal prediction in Figs.~\ref{fig:soccer}A and~B were similar, possibly because the soccer pitch ($105\,\mathrm{m} \times 68\,\mathrm{m}$) was larger than the basketball half-court ($14\,\mathrm{m} \times 15\,\mathrm{m}$). 
For the observation model, the results for a normal prediction in Fig.~\ref{fig:soccer}C indicate that defender~1 observed specific players and the ball.

\end{document}